\pdfoutput=1

\documentclass[11pt]{article}

\usepackage[preprint]{acl}

\usepackage{times}
\usepackage{latexsym}

\usepackage[T1]{fontenc}

\usepackage[utf8]{inputenc}

\usepackage{microtype}

\usepackage{inconsolata}

\usepackage{graphicx}

\usepackage{microtype}
\usepackage{hyperref}
\usepackage{url}
\usepackage{booktabs}
\usepackage{xcolor}
\usepackage{lineno}
\usepackage{amsmath}
\usepackage{enumitem}
\usepackage{graphicx,xparse}
\usepackage{makecell}
\usepackage{subcaption}
\usepackage{wrapfig}
\usepackage{tcolorbox}
\usepackage{multirow}
\usepackage{todonotes}
\usepackage{siunitx}
\usepackage{mathtools}
\usepackage[table]{xcolor}
\usepackage{tabularx}
\usepackage[dvipsnames]{xcolor}
\usepackage{xspace}
\NewDocumentCommand\emojione{}{\raisebox{-0.45em}{\includegraphics[height=1.25em]{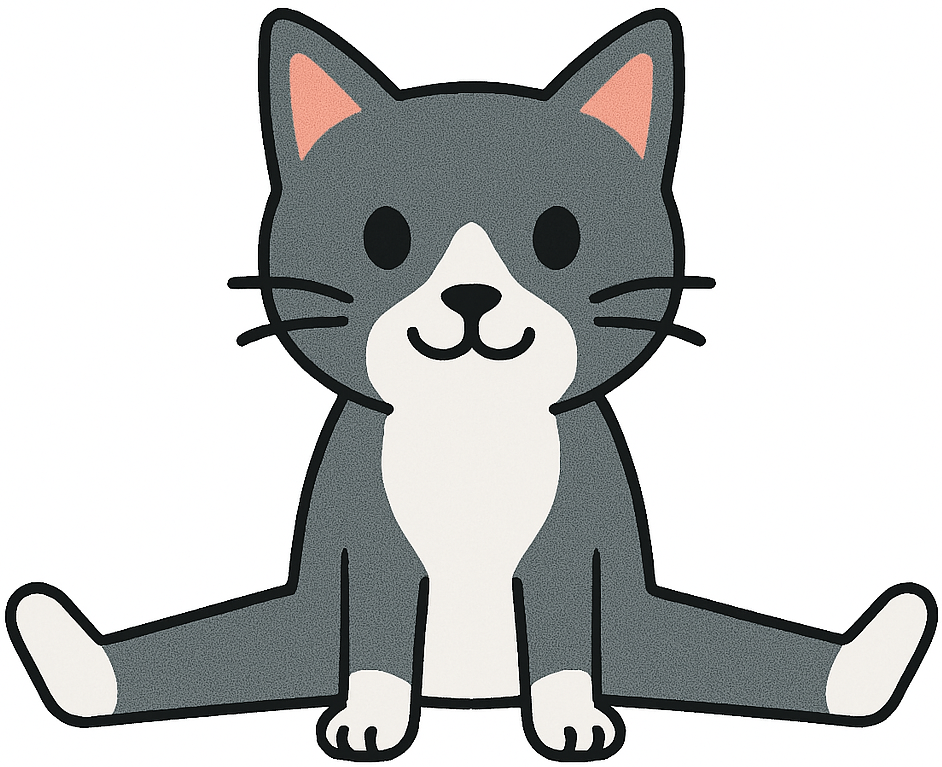}}}

\definecolor{darkpurple}{HTML}{4B0082} 
\newcommand{\splits}{\textsc{\textbf{\textcolor{darkpurple}{Splits!}}}\xspace}

\definecolor{darkblue}{rgb}{0, 0, 0.5}
\hypersetup{colorlinks=true, citecolor=darkblue, linkcolor=darkblue, urlcolor=darkblue}

\title{\emojione\hspace{0.1em} \splits Flexible Sociocultural Linguistic Investigation at Scale}

\author{Eylon Caplan, 
    Tania Chakraborty, \&      Dan Goldwasser \\
Department of Computer Science\\
Purdue University\\
West Lafayette, IN, USA \\
\texttt{\{ecaplan,tchakrab,dgoldwas\}@purdue.edu} 
}

\begin{document}
\maketitle

\begin{abstract}
Variation in language use, shaped by speakers' sociocultural background and specific context of use, offers a rich lens into cultural perspectives, values, and opinions. 
For example, Chinese students discuss \emph{healthy eating} with words like \emph{timing}, \emph{regularity}, and \emph{digestion}, whereas Americans use vocabulary like \emph{balancing food groups} and \emph{avoiding fat and sugar}, reflecting distinct cultural models of nutrition \cite{banna_cross-cultural_2016}. The computational study of these Sociocultural Linguistic Phenomena (SLP) has traditionally been done in NLP via tailored analyses of specific groups or topics, requiring specialized data collection and experimental operationalization---a process not well-suited to quick hypothesis exploration and prototyping. To address this, we propose constructing a ``sandbox'' designed for systematic and flexible sociolinguistic research.
Using our method, we construct a demographically/topically split Reddit dataset, \splits, validated by self-identification and by replicating several known SLPs from existing literature.
We showcase the sandbox's utility with a scalable, two-stage process that filters large collections of \emph{potential} SLPs (PSLPs) to surface the most promising candidates for deeper, qualitative investigation.\footnote{We release our \href{https://github.com/eyloncaplan/splits}{code}, our \href{https://huggingface.co/datasets/ecaplan/splits}{data}, and a \href{https://huggingface.co/spaces/ecaplan/splits}{sandbox demo}.}
\end{abstract}

\section{Introduction}

\begin{figure}[!t]
  \centering
  \includegraphics[width=\linewidth]{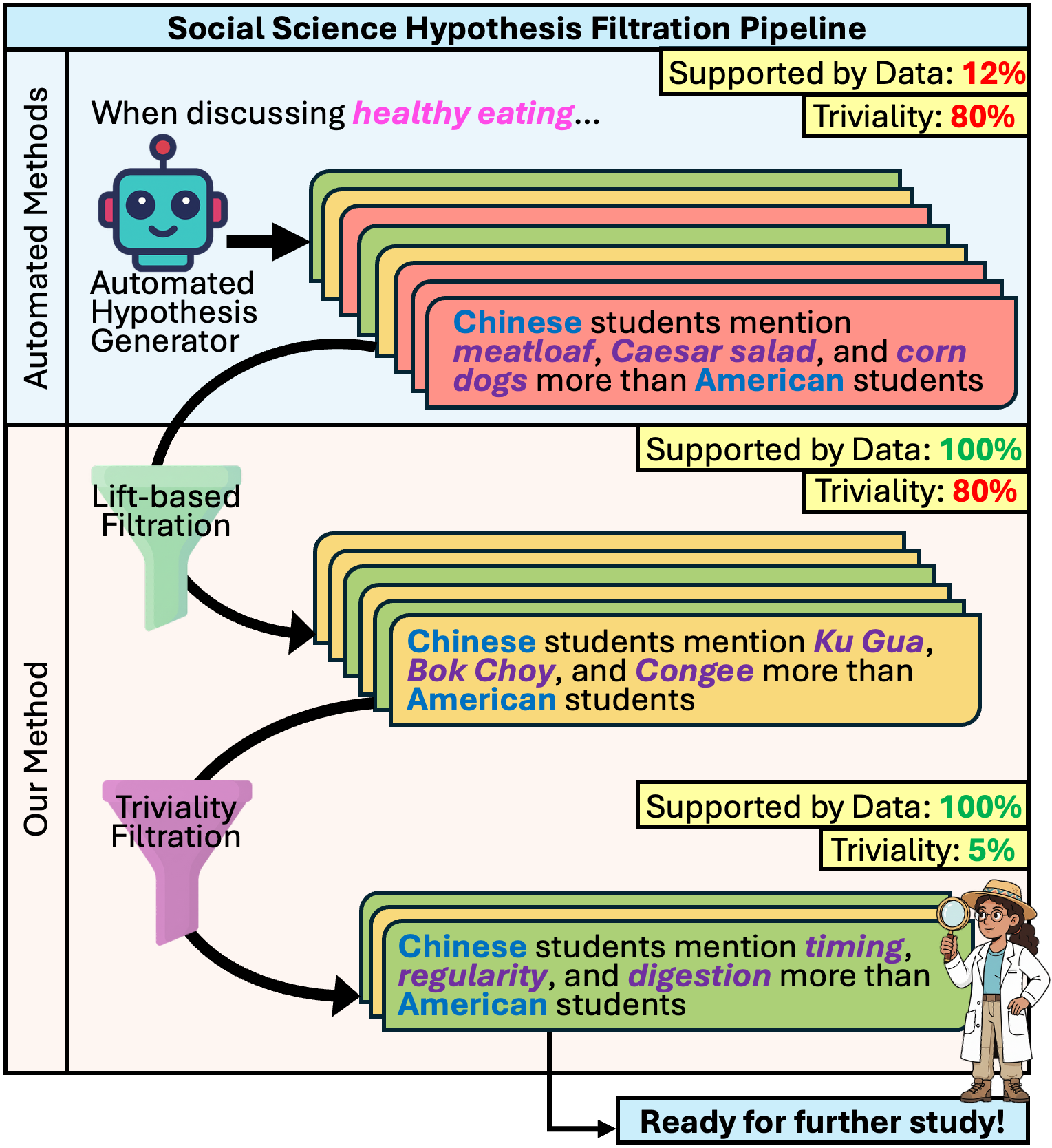}
  \caption{Automated methods can generate a large pool of hypotheses. Our goal is to surface the ones that are likely to be true (\textit{supported by data}) and interesting (\textit{non-trivial}). E.g., Chinese students mentioning Chinese food items is not surprising, but emphasis on \textit{timing} and \textit{regularity} in the context of ``healthy eating'' offers cultural insights.}
  \label{fig:funnel-diagram}
\end{figure}

How we speak is fundamental to who we are. Language is a powerful window into culture, allowing us to investigate the values and shared perspectives that shape a community's identity \cite{bucholtz_identity_2005}. Such indicative language manifests across many settings, in dialects with distinct grammar and vocabulary, like African American Vernacular English (AAVE) \cite{tia_african_2020, smitherman_african_2007}, but may also appear in more subtle, context-specific differences. For example, Figure~\ref{fig:funnel-diagram} shows an example of a \emph{sociocultural linguistic phenomenon} (SLP) wherein Chinese and American students discuss ``healthy eating'' with different vocabulary, reflecting how Americans are influenced by media/school nutrition messages, whereas Chinese students are influenced by Traditional Chinese Medicine \cite{banna_cross-cultural_2016}.\footnote{This work includes sensitive issues; discussion of demographics, representation, and misuse is in the Ethics Statement.}

Computational social science provides powerful tools to study specific SLPs within social media, yielding valuable insights into how different groups view and discuss contexts like politeness \cite{li_studying_2020, havaldar_comparing_2025}, values/moral foundations \cite{roy_tale_2023, borenstein_investigating_2024, caplan_conceptcarve_2025}, individualism \cite{havaldar_building_2024}, sustainability \cite{reuver_topic-specific_2024}, and health policy \cite{alliheibi_opinion_2021}. These are often tailored, deep dives of a single group/context, requiring significant effort in specialized data collection and experimental operationalization. Such methods are rigorous and powerful, but costly, making them difficult to extend to quick hypothesis exploration and idea prototyping.

The exhaustive and low-cost nature of a quick hypothesis exploration system would enable initial investigations of previously under-studied groups and contexts without requiring significant upfront investment. These investigations could identify promising phenomena worthy of deeper, bespoke study. To enable this rapid idea exploration, we envision a ``sandbox'' for systematically exploring the hypothesis space. Specifically, we seek to span the space of \emph{groups} that use language, and the space of \emph{contexts} in which language is used. 

To this end, we introduce a method for constructing such a ``sandbox'' using Reddit: with which we create the \splits dataset---split by both \textbf{user demographics} and \textbf{use contexts} operationalized as discussion topics (e.g., photography, travel, humor). We demonstrate its flexibility across various analytical approaches; for instance, we demonstrate one straightforward application by using \splits to confirm well-documented SLPs, such as \emph{code-switching} patterns of Black AAVE speakers. 

Recent work has focused on automatedly \emph{generating} scientific hypotheses, including social science hypotheses \cite{yang_large_2024, manning_automated_2024, peterson_using_2021}, yet a key bottleneck remains: how can a social scientist efficiently sift through thousands of computer-generated ideas to find the ones worth pursuing (Figure~\ref{fig:funnel-diagram})? As a second application of \splits, we address this by filtering for ``promising'' phenomena (worthy of further study) among a large pool of \emph{potential} SLPs (PSLPs).

To do this, we propose a scalable, two-step process (Figure \ref{fig:funnel-diagram}) that takes as input a large pool of candidate  PSLPs, and surfaces ones which are (1) supported by the data in the sandbox and (2) more likely to be unexpected, avoiding trivial PSLPs like ``Chinese students mention Chinese foods more than Americans''. Our contributions are as follows:
\begin{itemize}[leftmargin=*, nosep]\setlength\itemsep{0em}\setlength\topsep{0em}
    \item Propose a method for constructing a flexible, extensible ``sandbox'' for sociocultural linguistic investigation, and use it to create \splits, a 9.7 million-post dataset.
    \item Demonstrate the dataset's flexibility by reproducing known, literature-backed phenomena (AAVE code-switching and others).
    \item Propose a two-step process using \textit{lift} and \textit{triviality} to surface promising hypotheses from a large pool---intended for further expert investigation.
\end{itemize}

\section{Background and Related Work}
\subsection{SLPs and Variation}
We define a \emph{sociocultural linguistic phenomenon} (SLP) as any distinctive speech characteristic exhibited by a particular group. This concept is rooted in the field of sociolinguistics, which seeks to find ``correlations between social structure and linguistic structure'' \citep{gumperz1971language}. Specifically, our work connects to \textbf{variationist sociolinguistics}, which focuses on the social evaluation and use of linguistic variants through hypothesis-formation and statistical testing \citep{chambers2002studying}.

\begin{figure*}[!t]
    \centering
    \includegraphics[width=\linewidth]{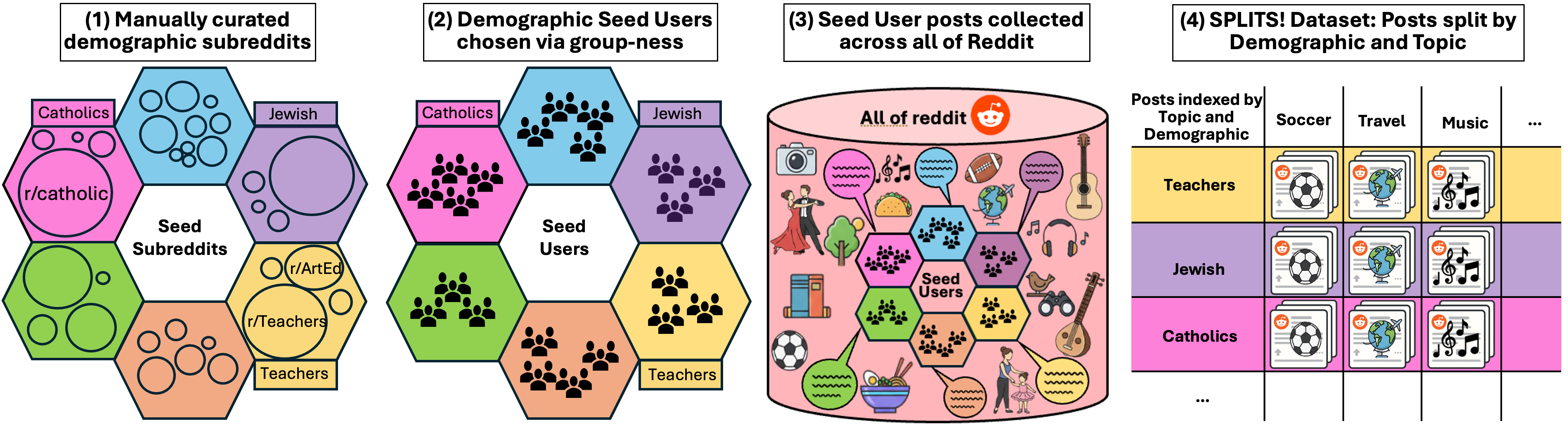}
    \caption{\splits \textbf{dataset creation:} (1) Set of seed subreddits are chosen for each demographic using manual inspection aided by statistical similarity metrics. (2) From all users in the seed subreddits, those with high "group-ness" are selected as seed users. (3) We collect all posts made by the seed users across all of reddit. (4) Posts are then labeled by content topic (independent of the source subreddit). The final dataset consists of these posts aggregated by demographic as well as topic.}
    \label{fig:dataset-creation-diagram}
\end{figure*}

Traditionally, variationist studies focus on how different linguistic forms are used to express the \textbf{same meaning} (e.g., phonetic variants of [r]) across different social groups \citep{wardhaugh_introduction_2021}. However, more recent ``third-wave'' approaches to variation have expanded this view to include \textbf{stancetaking}, where interlocutors use language to position themselves relative to the topic, each other, and broader social identities \citep{eckert2012three, jaffe2009sociolinguistic}. In this broader view, the fixed variable is not the meaning but the \textbf{context of use}, and the observed variation can be syntactic, lexical, or even semantic. Our definition of an SLP aligns with this broader conception: we hold the topic of discussion constant and analyze how different demographic groups vary their language to express different perspectives, priorities, and stances.

\subsection{Hypothesis Testing in Sociolinguistics}
Sociolinguistics is an empirical science, where the study of SLPs traditionally begins with \textbf{expert-led hypothesis formation}. Hypotheses are formed based on deep ethnographic knowledge gathered through methods like fieldwork and interviews \citep{wardhaugh_introduction_2021, hernandez-campoy_research_2014}. More recently, computational social science (CSS) has leveraged large-scale social media data to complement this process, enabling the generation and evaluation of social science hypotheses at a new scale \cite{yang_large_2024, manning_automated_2024, peterson_using_2021}. While the source of data (e.g., interview transcripts vs. online posts) and hypotheses (expert-derived vs. machine-generated) may differ, the core analytical step of using \textbf{correlational studies} and \textbf{statistical evaluation} to validate findings remains consistent.

\subsection{Lexica}
In this work, we represent SLPs as lexica. A \emph{lexicon} is a curated collection of words or phrases widely used for interpretable models in the social sciences \cite{hayati_does_2021, pryzant_deconfounded_2018, boyd2022development}. Lexica are a scalable and explainable method for analyzing large datasets \cite{havaldar_building_2024} and have been used to study sentiment, emotions, and mental health \cite{geng_inducing_2022}.
This approach is distinct from feature importance analysis (e.g., SHAP) \cite{ribeiro_why_2016, kim_interpretation_2020, lundberg_unified_2017}, which aims to interpret a model's predictions. Our focus, in contrast, is on using lexica to directly test hypotheses about the world as reflected in the data \cite{geng_inducing_2022, havaldar_building_2024}.

\subsection{Perspectivism and Social Media Corpora}
A significant body of sociolinguistically informed NLP research adopts a "Perspectivist" view: that language varies with a speaker's background and context \cite{slp-persp, lang-is-power, cult-aware-nlp, dialect-nlp, persp-survey}. This view has spurred the creation of perspective-aware datasets \cite{persp-dataset-design1, truth-is-lie} and models for tasks like stance detection \cite{stance1, stance2, stance3}. Many of these resources are social media corpora curated to study identity, often for tasks like demographic inference or bias detection \citep{sachdeva-etal-2022-targeted, wood-doughty-etal-2021-using, sap2020socialbiasframesreasoning, nadeem2020stereosetmeasuringstereotypicalbias, preotiuc-pietro-ungar-2018-user, tigunova_reddust_2020}. Our work builds on this tradition but differs in two key ways: our focus is on creating data splits by both demographic group and topic, and we do not attempt to attribute demographic authorship based on linguistic content.

\section{\splits Dataset Creation Method}
In this section, we present our method for constructing a Reddit ``sandbox'' for systematic sociolinguistic research. For the systematic approach, we ground our dataset's design in the two axes of variation described by \citet{maclagan_regional_2005}:
\begin{itemize}[leftmargin=*, nosep]
\item \textbf{User-level factors}: a speaker's long-term attributes (ethnicity, occupation, religion). We instantiate this axis using user \textbf{Demographic}.
\item \textbf{Use-level factors}: the context of an utterance (setting, purpose). We instantiate this axis by splitting our data by discussion \textbf{Topic}.
\end{itemize}
As such, we define a \textbf{demographic} as a collection of people with a shared, long-term user-level quality. We construct our dataset, \splits, from a corpus of Reddit data spanning 2012--2018 \cite{convokit}, collecting the top 50k subreddits by size. The final dataset contains 9.7 million posts, split across the 6 demographics and 89 topics.

\subsection{Curating Demographic subreddits}
In order to build the dataset, it was crucial to obtain posts that were truly written by some demographic group. To do this, we first identified subreddits that would be \emph{almost exclusively} used by people of some demographic, a set of subreddits which we refer to as the \emph{seed set}, shown in the first panel of Figure \ref{fig:dataset-creation-diagram}. For example, for the Catholic demographic, we included the subreddit r/CatholicDating, as it is likely to contain a very high proportion of Catholics. We define a \emph{user} of a subreddit as a Reddit account which at some point commented or posted in that subreddit. We began with a single high-quality seed subreddit and iteratively expanded this set by computing the user-overlap similarities (Jaccard, Cosine) to find related subreddits, which were then manually reviewed for inclusion. This process continued until no new relevant subreddits were found (visualized in Figure~\ref{fig:bubble-plot}, App.~\ref{app:dataset-details}).

We manually crafted 6 demographic seed sets: 2 ethnicities (African American/Black, Jewish descent), 2 occupations (teacher, construction worker), and 2 religious groups (Catholic, Hindu/Sikh/Jain). These demographics were chosen based on having relatively clean seed sets and being within an order of magnitude in total number of users. Broader discussion of demographic choices, including dimensions like gender and age are in App.~\ref{app:dataset-details}, with exact user and post counts.

\subsection{Demographic Seed Users via Group-ness}
The next step, as shown in part 2 of Figure \ref{fig:dataset-creation-diagram} was to choose the \emph{seed users} for each demographic. For each demographic, taking all posts in the union of the seed subreddits gave a seed demographic corpus $SD$, e.g. $SD_{\text{catholic}}$. We began by taking all unique users of the posts in $SD_{\text{demographic}}$, and got a demographic user set $U$, e.g. $U_{\text{catholic}}$ for $SD_{\text{catholic}}$. To make sure that the users of these posts truly belonged to the demographic, we devised a metric to measure a user's likely `group-ness'. The final seed user set for each demographic only consists of users that pass the group-ness threshold. The intuition was that a user is more likely to be in the target demographic if (1) they have many posts in the demographic seed set and (2) their posts are spread out across several subreddits (which we viewed as used \emph{almost exclusively} by the target demographic). We do this with the metric in Eq.~\ref{eq:metric}, in which we reward the total amount of activity \textit{and} diversity across subreddits in the seed set. 

\begin{equation}
\text{group-ness}(u) = \sum_{s \in S_D} \log(1 + c_{u,s})
\label{eq:metric}
\end{equation}

Here $u$ is a user, $S_D$ is the set of seed subreddits for the demographic $D$, and $c_{u,s}$ is the count of posts by user $u$ in subreddit $s$. We hypothesize that users with higher group-ness metric scores are likelier to be members of the target demographic. This is further validated in the next subsections.

\subsection{Collecting Seed User Posts Across Reddit}
Once we had the set of seed users for each demographic, the next step was to gather all the posts by the users on a variety of topics (Fig \ref{fig:dataset-creation-diagram}, part 3). We tracked the seed users across Reddit and collected all of their posts among \textbf{all subreddits} (not just the seed set). This yielded a set of posts $C$ for each demographic, where $SD_i \subset C_i$ for each group $i$.

\paragraph{Validating group-ness.} We wanted to ensure our group-ness metric guaranteed that the authors of the posts for each $C_i$ truly belonged to the demographic $i$. To test this, we created a set of self-identification phrases (e.g. for Catholic, ``I am a Catholic'', ``I'm Catholic'', etc.), and anti-self-identification phrases (e.g. ``I'm not a Catholic'', etc. and also ``I'm a Baptist'', ``I'm Jewish'', etc.). Some phrases can be found in Appendix~\ref{app:dataset-details}. Next, we searched for these phrases among all posts in a demographic's $C$, and to avoid false positives (e.g. quoted phrases, sarcasm, irony), we used an LLM to verify the context of these phrases (see prompt in App.~\ref{app:prompts}, human validation in App.~\ref{app:human-val-self-id}). Finally, we checked to see whether the group-ness metric correlated with higher self-identification rates, and whether it negatively correlated with anti-self-identification rates. To account for users with extremely high post count (chattiness) self-identifying very often, we normalized the self-identification rate by the average chattiness.

\begin{figure}[t]
    \centering
    \includegraphics[width=\columnwidth]{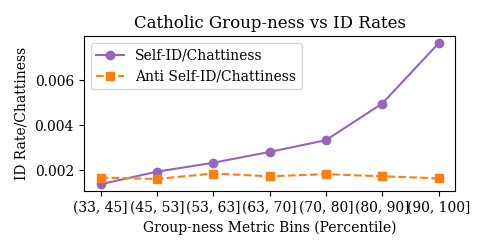}
    \caption{Does increasing group-ness isolate target demographic users? We plot self-identification rates vs. group-ness of the Catholic demographic, showing that it does.}
    \label{fig:duck-plots}
\end{figure}

Figure~\ref{fig:duck-plots} shows the result for the Catholic demographic. Similar successful validation trends were observed for all six retained demographics (see Appendix~\ref{app:dataset-details} for full plots). We note that we attempted a 7th demographic (Korean) which failed this validation due to insufficiently exclusive seed communities, highlighting a limitation of our methodology (see App.~\ref{app:korean-failure}). For the successful groups, past some threshold (usually about the 75th percentile in the group-ness metric), as a user's group-ness metric increases, their likelihood of self-identifying increases, and their likelihood of anti-self-identifying (i.e. saying they \emph{aren't} in the target demographic) goes down or stays the same. These thresholds of group-ness are chosen heuristically, as they reflect an inherent, necessary trade-off between demographic purity (precision) and dataset volume (recall). We select these manually based on empirical separation, and the exact threshold values are reported in our released code.

We make the following assumptions: (1) Redditors are generally honest when self-identifying, (2) a user's probability of explicitly self-identifying or anti-self-identifying in a given post is independent of their total posting frequency, and (3) honestly stating a self-identification or anti-self-identification phrase typically indicates genuine membership or non-membership. With these assumptions, the increasing separation we see between self-identification and anti-self-identification must mean that \textbf{users with a higher group-ness metric level are likelier to be in the target demographic}. That is, past some threshold, the datasets are `cleanly' posted by the target demographic. 

This approach inherently focuses our analysis on users \textbf{highly engaged in identity-centric subreddits}; while this creates a known selection bias, it provides a robust, high-confidence proxy for studying the language of these active online communities, a trade-off further discussed in Limitations.

\paragraph{Intersectionality.} As with all identities, the demographics we have selected may overlap in any individual. In practice, our dataset consists of demographics selected primarily for volume and exclusivity of subreddits. While some intersections are nearly nonexistent due to mutually excluding definitions (e.g. Catholic and Hindu/Jain/Sikh), others could have nontrivial intersections. 

To assess the empirical intersectionality of the selected demographic groups, we computed pairwise Jaccard similarity between sets of users in each group (full heatmap in App.~\ref{app:dataset-details}). Given the minimal empirical overlap observed (the greatest Jaccard was 0.0081), we chose to analyze the demographic groups separately in this work. We stress that this is a consequence of our sampling methodology and not a claim about the non-existence of intersectional identities in the real world. Acknowledging this limitation, we leave the computational study of intersectional SLPs as critical future work.

\subsection{Topics}
\label{sec:topics}

The final step was to label the posts for their topics. To create topic splits, we began with 11 categories (e.g., `Sports', `Entertainment'), using an LLM to generate about 20 specific topics (e.g., `basketball', `sci-fi') and corresponding query keywords for each (App.~\ref{app:dataset-details}). For each demographic's post collection ($C_i$), we filtered for users with high `group-ness' scores, setting the threshold for each group based on the observed separation between self- and anti-identification rates (Fig.~\ref{fig:duck-plots}). We then used the keywords to retrieve documents from these users' posts with the ColBERT retrieval model \cite{khattab_colbert_2020}. Finally, we used an LLM-based system to assess the topic relevance of each post (prompt in App.~\ref{app:prompts}, human validation in App.~\ref{app:human-val-topic}). We removed topics with too few posts after this step. The final dataset contains 9.7 million posts across 89 topics. We denote the set of posts for demographic $d$ on topic $t$ as $C_{d, t}$. A visual of the final \splits dataset is shown in part 4 of Fig \ref{fig:dataset-creation-diagram}.

\subsection{Replicating Case Studies with \splits}
\label{sec:case-studies}
To validate the richness and demonstrate one method of using \splits for answering sociocultural linguistic questions, we analyzed 5 literature-backed SLPs and show they are captured. For brevity, detailed explanations of each phenomenon, including literature and lexica are in App.~\ref{app:case-studies}.

\begin{table}[t]
\centering
\scriptsize
\setlength{\tabcolsep}{4pt}
\begin{tabular}{@{}l rr rr@{}}
\toprule
& \multicolumn{2}{c}{\textbf{Post Count}} & \multicolumn{2}{c}{\textbf{AAVE Use (\% posts)}} \\
\cmidrule(lr){2-3} \cmidrule(lr){4-5}
\textbf{Topic} & \textbf{Black} & \textbf{Non-Black} & \textbf{Black} & \textbf{Non-Black} \\
\midrule
Hip-Hop       & 56k & 18k  & 3.16\rlap{$^{*\dagger}$} & 2.00$^*$ \\
Professional  & 101k & 673k & 0.33\rlap{$^{\dagger}$} & 0.23 \\
\midrule
& \multicolumn{2}{c}{\textbf{Post Count}} & \multicolumn{2}{c}{\textbf{Yiddish Use (\% posts)}} \\
\cmidrule(lr){2-3} \cmidrule(lr){4-5}
\textbf{Topic} & \textbf{Jewish} & \textbf{Non-Jewish} & \textbf{Jewish} & \textbf{Non-Jewish} \\
\midrule
Judaism       & 97k & 64k & 0.19\rlap{$^{*\dagger}$} & 0.07$^*$ \\
Professional  & 135k & 639k & 0.01\rlap{$^{\dagger}$} & 0.01 \\
\midrule
& \multicolumn{2}{c}{\textbf{Post Count}} & \multicolumn{2}{c}{\textbf{`Dance' Use (\% posts)}} \\
\cmidrule(lr){2-3} \cmidrule(lr){4-5}
\textbf{Topic} & \textbf{\makecell{Hindu/\\Sikh/Jain}} & \textbf{\makecell{Non-Hindu/\\Sikh/Jain}} & \textbf{\makecell{Hindu/\\Sikh/Jain}} & \textbf{\makecell{Non-Hindu/\\Sikh/Jain}} \\
\midrule
\makecell{Personal\\Cultural Identity} & 88k & 381k & 0.44\rlap{$^{*}$} & 0.36 \\
\bottomrule
\end{tabular}
\caption{Can \splits replicate five known SLPs using lexical proportions? We show lexicon usage by demographic; $^{*}$denotes p-value $< 0.001$, between the two demographics (within topic).
    $^{\dagger}$denotes significance for the primary demographic's code-switching across topics. We show that all five SLPs are replicated.
}
\label{tab:case_studies_full}
\end{table}

Table~\ref{tab:case_studies_full} shows the results of using \splits to count post proportions that use lexica from existing literature. Analysis with our dataset replicates:
\begin{itemize}[leftmargin=*, nosep]\setlength\itemsep{0em}\setlength\topsep{0em}
    \item AAVE\footnote{African American Vernacular English}: Black users use AAVE features more frequently than non-Black users
    \item AAVE code-switching: Black users themselves use AAVE more when they discuss `Hip-Hop' than `Professional' topics
    \item Yiddish: Jewish users use Yiddish terms more frequently than non-Jewish users
    \item Yiddish code-switching: Jewish users themselves use Yiddish terms more when they discuss `Judaism' than `Professional' topics
    \item `Dance' as Identity in South Asians: when discussing `Personal Cultural Identity', Hindus/Jains/Sikhs mention `dance' and `dancing' more than non-group members
\end{itemize}

\begin{figure}[tb]
  \centering
  \footnotesize
  \includegraphics[width=\linewidth]{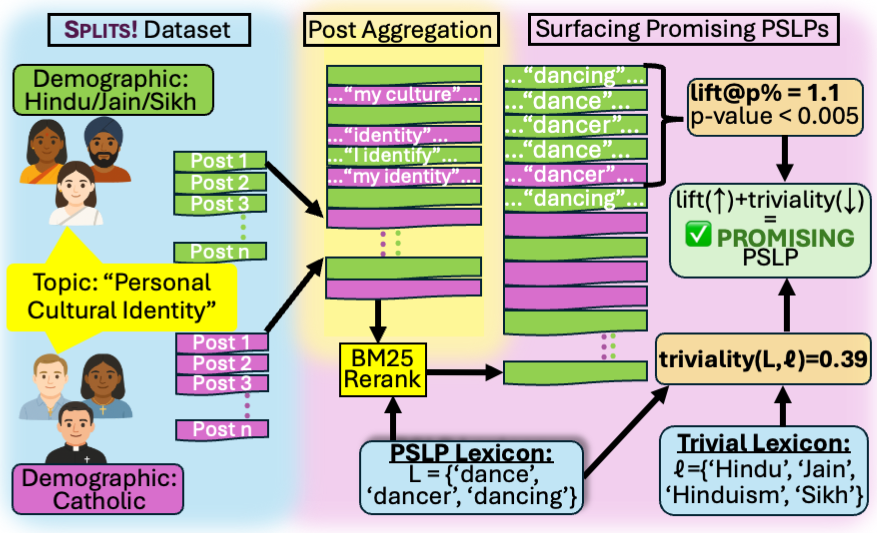}
  \caption{Main steps to determine if a PSLP is \textit{promising} using \splits: (1) Aggregate posts from 2 demographics about the same topic (2) Use the PSLP's lexicon to rerank the aggregated posts (3) Compute \texttt{lift} and \texttt{triviality} for the PSLP (4) Promising PSLPs achieve high lift \& low triviality.}
  \label{fig:splits-eval-diagram}
\end{figure}

\section{Two-Stage PSLP Filtration Process}

This section details another method of using \splits: quickly filtering for PSLPs' (1) alignment with data and (2) non-triviality---worthiness for future study. The method relies on the \emph{specific class} of PSLPs testable by this lexical form: ``demographic $A$ uses set $L$ of words and phrases more than demographic $B$ when discussing topic $t$''. We represent this as PSLP$_{L, A, B, t}$.

Figure~\ref{fig:splits-eval-diagram} shows the high level: we propose pairing the corpora of 2 demographics discussing the \emph{same} topic, indexing them together, and measuring how much better than random a lexicon is able to rerank the set. This method is motivated by the \emph{speed} and \emph{reusability} conferred by indexing for rapid testing of PSLPs. We stress that \textbf{not all SLPs can be put in this form}; our pipeline is intentionally scoped to test lexical-level hypotheses for tractability and interpretability.

\subsection{Quantifying Validity with Lift} To ensure enough posts for each topic, topic/demographic post sets $C_{d, t}$ with fewer than 2,000 posts were dropped. Then, for every topic $t$, and two demographics $A, B$, the two post sets $C_{A, t}$ and $C_{B, t}$ were combined and indexed together using the pyserini BM25 implementation \cite{Lin_etal_SIGIR2021_Pyserini}. A lexical method (BM25) was chosen as a non-`black box' model, conferring interpretability compared to neural models. Hence there were 781 unique indices $I_{A, B, t}$ for every combination of demographics $A, B$ and topic $t$. App.~\ref{app:dataset-details} visualizes the pairings, the number of posts by demographic, and by topic category.

 Given a PSLP's lexicon $L$, we can then test its validity over our dataset by doing the following: Using the relevant index $I_{A, B, t}$ to the PSLP, we rerank the whole set using $L$ as a query and the phrase-aware BM25 algorithm (Figure~\ref{fig:splits-eval-diagram}). To measure $L$'s success in pulling demographic $A$ upward, we use the metric of \emph{lift} from Data Mining \cite{tuffery_data_2011}. We define lift@p\% of demographic $A$ in Eq.~\ref{eq:lift}.

\begin{equation}
    \small
    \text{lift@}p\% = \frac{\# A \text{ posts@}p\%/\#\text{ posts@}p\%}{\# A \text{ posts overall}/\#\text{ posts overall}}
    \label{eq:lift}
\end{equation}

Here $p\%$ is the top $p$ percent of posts in the ranking. Lift@p\% $> 1$ indicates that $L$ has pulled $A$ over $B$ more than random. To guarantee significance, we also perform a one-tailed hypergeometric test, which is the exact distribution of randomly selected top p\% posts. The magnitude of the lift indicates the strength of $L$ pulling up $A$ over $B$, and the p-value indicates the significance of the PSLP overall. Throughout this paper, we report lift@0.5\%, though we compute it for 1\%, 2\%, 5\%, and 10\% as well. Generally, smaller values of $p\%$ capture more rare/subtle phenomena, while higher values only capture very prominent phenomena.

\subsection{Quantifying Triviality} High significance lift is not enough to indicate a promising PSLP for future study. Despite each index being limited to a specific topic, any demographic is inherently more likely to use words that are `trivial' to that demographic. By this, we mean words that are definitional in nature (e.g. ``Jewish'', ``Jew'', ``Judaism'' for the Jewish demographic), or words that are exclusive/nearly exclusive to the demographic (e.g. ``Diwali'', ``kirpan'', ``ahimsa'' for the Hindu/Jain/Sikh demographic).

To filter out trivially lifting PSLPs, we propose a `triviality' metric. At a high level, this measures the similarity between the PSLP's lexicon $L$ and the target demographic $A$ (Figure~\ref{fig:splits-eval-diagram}). To operationalize this, we manually authored a small lexicon $\ell$ for each demographic, containing 5-10 terms (e.g. $\ell_{jewish}$ as ``"Jewish'', ``Jew'', ``Judaism'', ``Jewish holidays'', ...; see App.~\ref{app:pslp-automatic-evaluation} for full lists).

We use the subspace recall-like `$R_{subspace}$' score introduced in \citet{ishibashi_subspace_2024} to measure the similarity between two sets of words (utilizing bert-base-uncased embeddings), scored within $[0, 1]$. We precisely define triviality as 

\begin{equation}
triv(PSLP_{L, A, B, t}) \coloneqq R_{subspace}(L, \ell_{A}).
\end{equation}

As such, the more words in the lexicon $L$ that are semantically similar to the target demographic $A$ as a whole, the more trivial it becomes. We note that the `triviality' metric can be easily tuned to fit specific use cases. We acknowledge that with this definition, \emph{non}-triviality does not perfectly map onto ``interesting-ness'' or ``unexpectedness'', but in sections \S~\ref{sec:human-validation} and \S~\ref{sec:lit-llm-pslps}, we show that it correlates with human judgments of ``unexpectedness'', and the ``interesting-ness'' of research findings.

\section{Validation of Lift and Triviality}
\begin{figure}[tb]
    \centering
    \footnotesize
    \begin{subfigure}[b]{\linewidth}
        \centering
        \includegraphics[width=\linewidth]{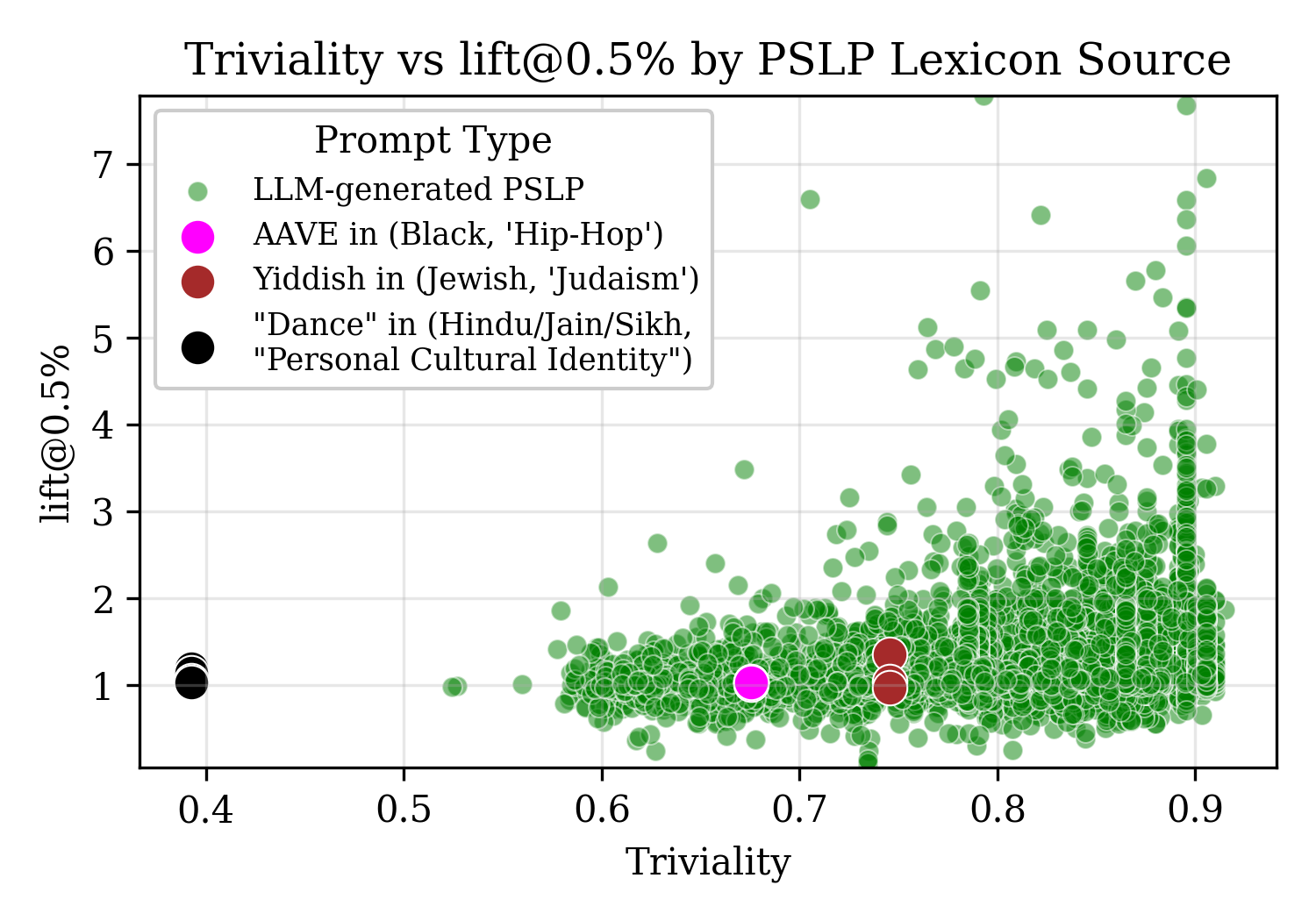}
    \end{subfigure}
    \caption{Are higher-lift PSLPs more trivial? We plot LLM PSLPs (subsampled) along with case studies. The upward trend indicates many obvious hypotheses achieve high lift.}
    \label{fig:PSLP-plot}
\end{figure}

We now motivate and validate our metrics.

\subsection{Why Statistical Significance is Insufficient}
To demonstrate our filtration process's ability to filter a large and noisy set of candidates, we used an LLM to generate over 23,000 PSLPs across all demographic-pair-topic combinations. We stress that \textbf{our goal was not to design a novel hypothesis generator}, but to produce a large, diverse set of candidate PSLPs for our proposed filtration process at scale. Further candidate details in Appendix~\ref{app:pslp-automatic-evaluation}.

Figure~\ref{fig:PSLP-plot} plots the resulting Lift versus Triviality score for each of these 23,000+ PSLPs, with our case studies from \S~\ref{sec:case-studies} included for reference (case study demographic vs. demographic lifts are in App.~\ref{app:case-studies}). The plot reveals the central motivation for our proposed filtration process: there is a significant positive correlation (0.32 Spearman for lift@0.5\%) between a PSLP's triviality and its statistical lift. This finding confirms that \textbf{relying on statistical significance alone is insufficient}, since \textbf{many commonsense or obvious hypotheses also achieve high lift}. This necessitates our process's second stage to isolate the more promising, non-trivial candidates for expert review.

\begin{table*}[tb]
    \centering
    \scriptsize
    
    \definecolor{hindugreen}{HTML}{A7D1A7}  
    \definecolor{jewishgreen}{HTML}{CFE5CF} 
    \definecolor{catholicred}{HTML}{FFCCCC}  
    
    \setlength{\tabcolsep}{2pt}
    \renewcommand{\arraystretch}{0.99}

    \begin{tabularx}{\textwidth}{@{} ll l >{\raggedright\ttfamily}X rr @{}}
        \toprule
        \textbf{Target} & \textbf{Contrast} & \textbf{Topic} & \textbf{Lexicon} & \multicolumn{1}{c}{\textbf{\begin{tabular}[c]{@{}c@{}}Triviality \\ (percentile) $\downarrow$\end{tabular}}} & \multicolumn{1}{c}{\textbf{\begin{tabular}[c]{@{}c@{}}Human \\ Score $\uparrow$ [1, 5]\end{tabular}}} \\
        \midrule
        
        \rowcolor{hindugreen}
        Hindu/Jain/Sikh & Teacher & Books/Literature & powerful woman, making her own choices, emotional complexity, ... & 8.80 & * \\
        
        \rowcolor{jewishgreen}
        Jewish & Catholic & Healthcare & early detection, preventative care, screening, proactive, ... & 12.10 & 4.67 \\
        
        \rowcolor{catholicred}
        Catholic & Black & Healthcare & papal authority, church teachings, bishop, doctrine, catechism, ... & 99.38 & 1.00 \\
        
        \bottomrule
    \end{tabularx}
    
    \caption{What do filtered PSLPs look like? Row 1: literature-inspired  PSLP \cite{jain_reinterpretation_2016} (interesting), Row 2: LLM-generated PSLP (non-trivial, unexpected), Row 3: LLM-generated PSLP (trivial, expected).}
    \label{tab:llm-lexica}

\end{table*}

\begin{figure}[t]
  \centering
  \includegraphics[width=\linewidth]{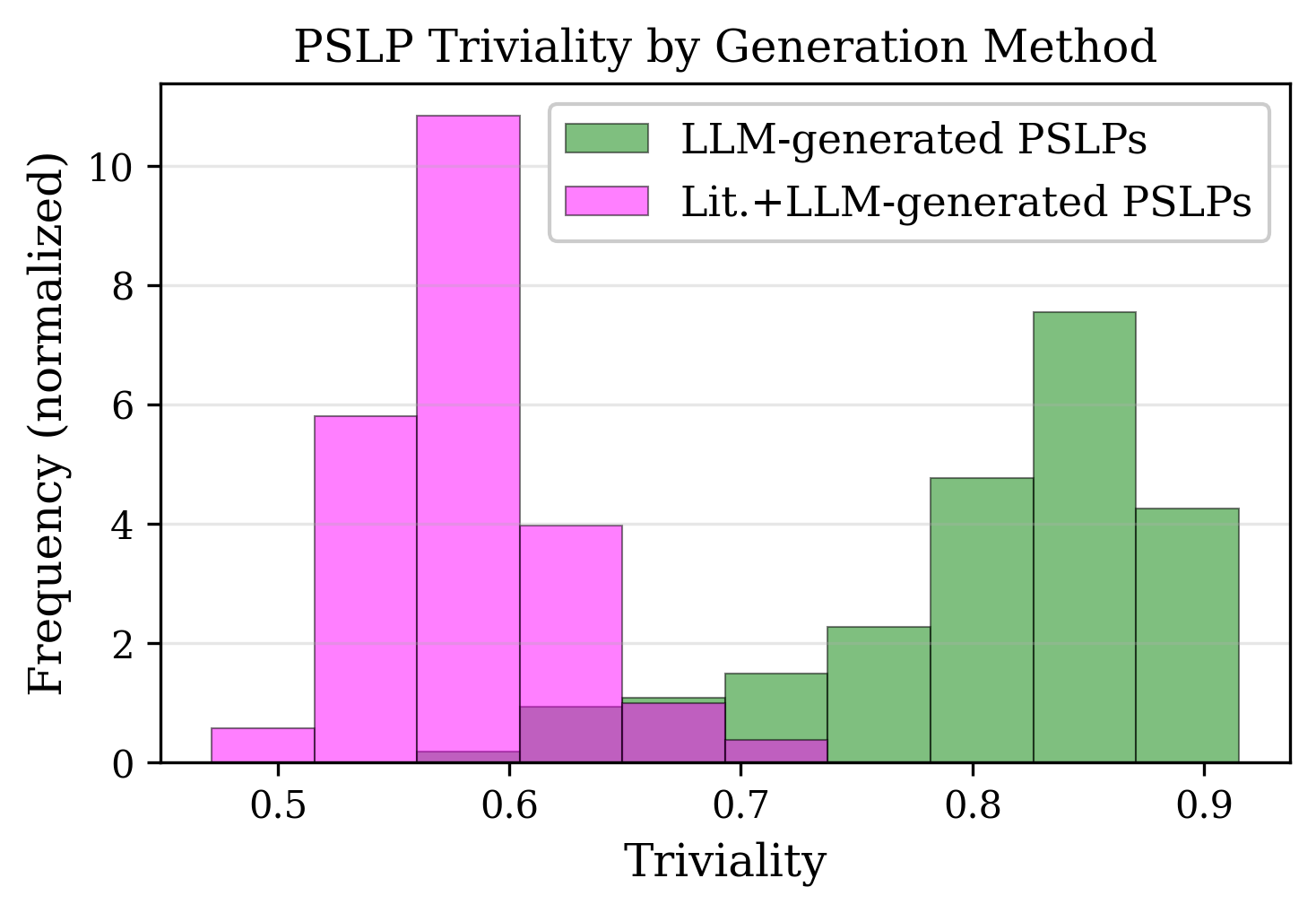}
  \caption{Are PSLPs inspired by social science findings less trivial? We plot the triviality of these PSLPs (pink) against normal LLM PSLPs (green), showing they are far less trivial.}
  \label{fig:llm-vs-llm-lit-triv}
\end{figure}

\subsection{Human Validation of the Triviality Metric}
\label{sec:human-validation}

To validate our automated Triviality metric, we measured its alignment with human judgments of ``unexpectedness.'' We conducted a study where 9 demographically diverse annotators, including members of 4 of the 6 groups in our dataset, rated 500 PSLPs. For each PSLP (a lexicon, a demographic pair, and a topic), they rated its unexpectedness on a 1-5 scale (5 being highly unexpected). The annotation task proved reliable, achieving an Intraclass Correlation Coefficient (ICC(2,k)) of 0.74, indicating good agreement. As hypothesized, we found a significant negative correlation (Spearman's $\rho = -0.38$) between our Triviality score and the average human score. This confirmed that our metric can be a useful (though imperfect) heuristic filter for prioritizing unexpected candidates.

\subsection{Validating Triviality Against Published Sociocultural Research}
\label{sec:lit-llm-pslps}

To further validate the Triviality metric, we test its ability to distinguish standard LLM-generated hypotheses from those grounded in existing academic literature. Our rationale is that \textit{findings published in peer-reviewed social science papers represent a reliable proxy for what domain experts consider `interesting'}. If our Triviality metric is effective, it should assign significantly lower scores to hypotheses derived from expert-vetted sources.

\paragraph{Literature-Inspired PSLPs.} To construct this test set, we manually curated 11 social science papers by searching Google Scholar for literature on specific demographic-topic intersections present in our dataset (e.g., `Teacher Anime/Manga', `Catholic Video Games') (see App.~\ref{app:lit-inspired-pslps}). Selected papers had findings that were directly relevant to the target demographic and topic. For each paper, we prompted an LLM to first summarize its core findings and then, based on that summary, generate 10-15 lexica that operationalize the paper's conclusions (prompt in App.~\ref{app:prompts}). The result was 132 lexica. A human validation study confirmed that these generated lexica faithfully reflect the findings of their source papers (see App.~\ref{app:human-val-lit-pslp}). To create testable PSLPs, we paired these lexica with all available contrast demographics in our dataset, yielding 591 total \textit{Literature-Inspired PSLPs}.

Figure~\ref{fig:llm-vs-llm-lit-triv} compares the Triviality score distributions for standard LLM-generated PSLPs vs. our Literature-Inspired PSLPs. The distinction is stark: \textbf{Literature-Inspired PSLPs exhibit a markedly lower distribution of Triviality scores} (mean of 0.585 vs. 0.810). This result demonstrates that our Triviality metric generally \textit{aligns with the concept of academic sociocultural `interestingness,'} serving as a valuable heuristic to prioritize hypotheses worthy of deeper investigation.

\section{Surfacing Promising PSLPs}
\label{sec:analysis}

\begin{table}[b]
  \centering
  \renewcommand{\arraystretch}{0.9}
  \setlength{\tabcolsep}{3pt}
  \resizebox{\columnwidth}{!}{
  \begin{tabular}{ccccccc}
    \toprule
    \textbf{\shortstack{Percentile\\threshold}} & \textbf{Precision} & \textbf{Recall} & \textbf{F1} & \textbf{\shortstack{Effort\\(\# inspected/\\promising)}} & \textbf{Speed-up} & \textbf{$p$-value} \\
    \midrule
    1 (Baseline) & \textbf{0.270} & 1.000 & 0.425 & $3.70$ & $1.00\times$ & -- \\
    \midrule
    0.1 & \textbf{0.480} & 0.178 & 0.259 & $2.08$ & $1.78\times$ & 0.001 \\
    0.2 & \textbf{0.455} & 0.341 & 0.390 & $2.20$ & $1.68\times$ & $<10^{-3}$ \\
    0.3 & \textbf{0.447} & 0.496 & 0.470 & $2.24$ & $1.65\times$ & $<10^{-3}$ \\
    0.4 & \textbf{0.425} & 0.630 & 0.507 & $2.35$ & $1.57\times$ & $<10^{-3}$ \\
    0.5 & \textbf{0.398} & 0.741 & 0.518 & $2.51$ & $1.47\times$ & $<10^{-3}$ \\
    \bottomrule
  \end{tabular}
  }
  \caption{Triviality used for `unexpectedness' classification of LLM PSLPs. The Percentile Threshold is the Triviality cut-off for classifying ``unexpected''.}
  \label{tab:performance-metrics}
\end{table}

We now apply our pipeline on a large pool of LLM-generated hypotheses. We quantify the efficiency of our two-step filtration and then present a qualitative analysis of some surfaced PSLPs.

\subsection{Quantitative Analysis of Filtration}
Our pipeline is designed to help researchers efficiently navigate vast hypothesis spaces. To demonstrate this, we applied our two-step filtration process to the over 23,000 PSLPs generated by an LLM.
First, we filtered for statistical significance (p < 0.025), which by itself \textbf{reduced the candidate pool by over 10x}, identifying $\sim$2,300 PSLPs that are supported by the data. The subsequent challenge is to find the truly ``promising'' candidates within this set—those that are not only supported by data but also deemed unexpected by human experts (avg. score > 3/5).

This is where our Triviality metric provides a crucial second filter. By using it to prioritize the inspection of the $\sim$2,300 supported candidates, we reduce the manual ``effort'' (number of PSLPs inspected per promising find) by \textbf{an additional 1.5-1.8x} (Table \ref{tab:performance-metrics}). In this particular LLM-based generation scenario, the full two-step process provides a \textbf{combined 15-18x reduction in effort}, showcasing its value in helping researchers quickly isolate a high-potential set of hypotheses from a large, noisy initial pool for deeper qualitative study.

\subsection{Qualitative Analysis of Promising PSLPs} 

Table \ref{tab:llm-lexica} showcases significantly lifting PSLPs surfaced by our pipeline, illustrating the distinction between those deemed promising and those filtered out as trivial. A particularly compelling, non-trivial example suggests that Jewish users, when discussing healthcare, employ a lexicon of `preventative care', `early detection', and being `proactive' significantly more than Catholic users.

This linguistic pattern may point to deeper, culturally-ingrained perspectives and can provide a concrete, data-driven \textbf{starting point} for social scientists to further explore this space. For example, theological scholarship contrasts Judaism’s focus on the present world with Christianity's historical emphasis on the afterlife \cite{mcdermott_thumbnail_2015, schwartz_is_2011, eckardt_death_1972}.  We stress again that PSLPs are only \emph{potential} SLPs---a \textbf{first step} for further study, not foregone conclusions.

\section{Conclusion}
We presented a method for building a flexible sociolinguistic ``sandbox'' for the rapid exploration of cultural language use, creating the \splits dataset as an instance. We demonstrated its utility through a two-stage filtration process that uses \textit{lift} and \textit{triviality}. This approach effectively narrows a vast space of computer-generated hypotheses to a manageable set of promising candidates, bridging the gap between large-scale computational analysis and nuanced, expert-led sociolinguistic inquiry.

Future work could attempt to capture richer linguistic dimensions beyond lexica, such as syntactic structures, semantic framing, or pragmatic features. However, transitioning to more semantically complex or neural representations must be approached cautiously; allowing for non-lexical features introduces the risk of uninterpretable model bias.

\section*{Ethics Statement}

Research into language and its relation to identity carries significant ethical responsibilities. We have sought to address these responsibilities throughout our research design, dataset curation, and framework development.

\subsection*{Risk of Stereotyping and Social Essentialism}
This paper investigates linguistic variation across demographic groups. Any such work risks engaging in or enabling harmful stereotyping. We are particularly concerned with avoiding Social Essentialism—the fallacious belief that social groups possess distinct, inherent essences \cite{define-essentialism, essentialism}. In AI, this can manifest as systems that unfairly generalize or reinforce societal biases \cite{ess-is-bad}.

Our methodology is explicitly designed to counter this risk. Following recent work in cognitive science and NLP that advocates for more nuanced models of identity \cite{influence-on-ling, towards-countering-ess}, the \splits dataset is structured not only by demographic but also by 89 distinct topics of discussion. This design is rooted in our guiding principle: \textbf{language is expressed and interpreted in nuanced ways that extend well beyond a person's demographic membership}. By enabling analysis that is always conditioned on a specific context, our framework encourages a perspectivist view of language and has the potential to reveal nuanced differences that actively counter broad, essentialist stereotypes.

\subsection*{Data Source and Privacy}
This work is built on publicly available Reddit posts sourced from the Convokit project \cite{convokit}. While the data is public, we acknowledge that users may not anticipate their posts being used in academic research. To mitigate privacy risks, the public release of \splits does not contain Reddit usernames. To enable user-level analysis, we provide pseudonymized user IDs.

\subsection*{Intended Use and Prohibited Misuse}
We provide this dataset and framework with clear guidelines on its intended and prohibited uses.

Intended Use: \splits is designed for non-commercial, academic research into observational sociocultural linguistics. Its purpose is to help researchers explore how language use varies across different contexts and to generate hypotheses for further qualitative or quantitative investigation.

Prohibited Misuse: This dataset is not suitable for any application that assigns scores, makes judgments about, or could otherwise stereotype individuals based on their perceived group membership.

\section*{Limitations}
We list the limitations of our work here.

\paragraph{Representational Bias.} The data is sourced from Reddit and is not representative of the global population or even the full population of the demographic groups studied. Therefore it inherently contains {selection bias}.

The dataset reflects the language of English-speaking, active Reddit users within these communities. It cannot capture the full richness or complexity of their lived experiences and should not be interpreted as doing so.

\paragraph{Temporal Bias.} The data spans from 2012 to 2018. Language, cultural norms, and the topics of online discourse evolve rapidly. This dataset provides a valuable historical snapshot, but the patterns observed may not reflect the contemporary linguistic practices or views of these communities.

\paragraph{Platform-Specific Discourse.} Reddit has a unique culture with its own jargon, memes, and conversational norms. The linguistic phenomena identified may be intertwined with the specific communicative conventions of the Reddit platform and may not generalize to other social media contexts or offline conversations.

\paragraph{Demographic Labeling and Intersectionality.} Our demographic labels are high-confidence proxies derived from user activity in ``seed'' subreddits, not from direct self-declaration. This method may select for users who are particularly active and vocal in identity-centric communities. A key risk of this approach is that \textbf{it may amplify perceived linguistic differences between groups}, as our sample may not be representative of the broader demographic, but rather a subset with stronger in-group identification. Furthermore, our broad demographic categories (e.g., ``Black,'' ``Catholic'') do not capture the immense intra-group diversity. Our current framework treats these groups as discrete and is not designed to analyze intersectional identities, where multiple identity facets jointly shape language use. The computational study of such intersectional phenomena is a critical direction for future work.

\paragraph{Seed and $\ell$ Selection.} Our methodology relies on two main manual decisions: demographic seed subreddits, and `trivial' demographic lexica. For the demographic seed sets, we relied on subreddit names and descriptions. We believe this process is transparent and replicable by any informed researcher without requiring privileged ``insider'' knowledge of a group. Similarly, the Triviality lexica ($\ell_A$) were constructed using terms that are fundamentally definitional to each demographic, ensuring broad agreement and minimizing subjectivity.

\paragraph{LLM-based Topic Creation.} Our topic-splitting process uses LLMs for keyword generation and relevance filtering. While powerful, these models are not infallible and may introduce noise or systematic biases into the topic definitions, potentially affecting the validity of cross-topic comparisons.

\paragraph{PSLP Representation.} Our framework operationalizes PSLPs as lexica of words and phrases. This was a deliberate design choice to ensure scalability and interpretability. However, this simplification cannot capture more complex phenomena rooted in deeper semantic or pragmatic nuance, such as sarcasm, narrative structure, or connotative meaning. Our framework should therefore be seen as a tool for identifying lexical-level differences, which can serve as a starting point for more nuanced qualitative or computational investigation.

\paragraph{LLM-Generated Hypothesis Bias.} While our pipeline utilizes LLMs to rapidly generate a large pool of candidate PSLPs, researchers must remain highly critical of these models when generating hypotheses. LLMs inherently encode biases from their pre-training data and may hallucinate, reproduce harmful stereotypes, or over-represent majority viewpoints. Consequently, the hypotheses surfaced by our pipeline should never be taken as factual sociolinguistic claims, but rather treated strictly as exploratory starting points for rigorous human-led validation.

\paragraph{Embedding Bias in Triviality.} Our automated Triviality metric relies on BERT embeddings to measure the semantic similarity between a PSLP lexicon and a target demographic lexicon. Because pre-trained models contain inherent representational biases, these embeddings may possess certain blind spots. Consequently, the Triviality metric might misjudge or fail to capture the semantic nuance of culturally specific, historical, or marginalized concepts, potentially filtering out valid phenomena or missing hidden trivialities.

\section*{Acknowledgments}
We thank the anonymous reviewers for helping to strengthen the paper, and our annotators for generously volunteering their time.

\bibliography{custom}

@article{ess-is-bad,
  title        = {Towards Countering Essentialism through Social Bias Reasoning},
  author       = {Emily Allaway and Nina Taneja and Sarah‑Jane Leslie and Maarten Sap},
  year         = {2023},
  journal      = {arXiv preprint arXiv:2303.16173},
  url          = {https://arxiv.org/abs/2303.16173},
  note         = {Preprint}
}

@article{essentialism,
  title        = {Cultural transmission of social essentialism},
  author       = {Marjorie Rhodes and Sarah‑Jane Leslie and Christina M. Tworek},
  journal      = {Proceedings of the National Academy of Sciences},
  year         = {2012},
  volume       = {109},
  number       = {34},
  pages        = {13526--13531},
  doi          = {10.1073/pnas.1208951109}
}

@article{define-essentialism,
  title        = {Psychological essentialism in children},
  author       = {Susan A. Gelman},
  journal      = {Trends in Cognitive Sciences},
  year         = {2004},
  volume       = {8},
  number       = {9},
  pages        = {404--409},
  doi          = {10.1016/j.tics.2004.07.001}
}

@misc{dialect-nlp,
      title={Natural Language Processing for Dialects of a Language: A Survey}, 
      author={Aditya Joshi and Raj Dabre and Diptesh Kanojia and Zhuang Li and Haolan Zhan and Gholamreza Haffari and Doris Dippold},
      year={2024},
      eprint={2401.05632},
      archivePrefix={arXiv},
      primaryClass={cs.CL},
      url={https://arxiv.org/abs/2401.05632}, 
}

@article{cult-aware-nlp,
    title = "Culturally Aware and Adapted {NLP}: A Taxonomy and a Survey of the State of the Art",
    author = "Liu, Chen Cecilia  and
      Gurevych, Iryna  and
      Korhonen, Anna",
    journal = "Transactions of the Association for Computational Linguistics",
    volume = "13",
    year = "2025",
    address = "Cambridge, MA",
    publisher = "MIT Press",
    url = "https://aclanthology.org/2025.tacl-1.31/",
    doi = "10.1162/tacl_a_00760",
    pages = "652--689"
}

@misc{lang-is-power,
      title={Language (Technology) is Power: A Critical Survey of "Bias" in NLP}, 
      author={Su Lin Blodgett and Solon Barocas and Hal Daumé III and Hanna Wallach},
      year={2020},
      eprint={2005.14050},
      archivePrefix={arXiv},
      primaryClass={cs.CL},
      url={https://arxiv.org/abs/2005.14050}, 
}

@inproceedings{slp-persp,
    title = "On learning and representing social meaning in {NLP}: a sociolinguistic perspective",
    author = "Nguyen, Dong  and
      Rosseel, Laura  and
      Grieve, Jack",
    editor = "Toutanova, Kristina  and
      Rumshisky, Anna  and
      Zettlemoyer, Luke  and
      Hakkani-Tur, Dilek  and
      Beltagy, Iz  and
      Bethard, Steven  and
      Cotterell, Ryan  and
      Chakraborty, Tanmoy  and
      Zhou, Yichao",
    booktitle = "Proceedings of the 2021 Conference of the North American Chapter of the Association for Computational Linguistics: Human Language Technologies",
    month = jun,
    year = "2021",
    address = "Online",
    publisher = "Association for Computational Linguistics",
    url = "https://aclanthology.org/2021.naacl-main.50/",
    doi = "10.18653/v1/2021.naacl-main.50",
    pages = "603--612"
}

@misc{towards-countering-ess,
      title={Towards Countering Essentialism through Social Bias Reasoning}, 
      author={Emily Allaway and Nina Taneja and Sarah-Jane Leslie and Maarten Sap},
      year={2023},
      eprint={2303.16173},
      archivePrefix={arXiv},
      primaryClass={cs.CL},
      url={https://arxiv.org/abs/2303.16173}, 
}

@article{influence-on-ling,
  title     = {The Influence of Linguistic Form and Causal Explanations on the Development of Social Essentialism},
  author    = {Benitez, Jaclyn and Leshin, R. A. and Rhodes, Marjorie},
  journal   = {Cognition},
  volume    = {229},
  pages     = {105246},
  year      = {2022},
  doi       = {10.1016/j.cognition.2022.105246},
  url       = {https://doi.org/10.1016/j.cognition.2022.105246}
}

@inproceedings{persp-dataset-design1,
    title = "The Perspectivist Paradigm Shift: Assumptions and Challenges of Capturing Human Labels",
    author = "Fleisig, Eve  and
      Blodgett, Su Lin  and
      Klein, Dan  and
      Talat, Zeerak",
    editor = "Duh, Kevin  and
      Gomez, Helena  and
      Bethard, Steven",
    booktitle = "Proceedings of the 2024 Conference of the North American Chapter of the Association for Computational Linguistics: Human Language Technologies (Volume 1: Long Papers)",
    month = jun,
    year = "2024",
    address = "Mexico City, Mexico",
    publisher = "Association for Computational Linguistics",
    url = "https://aclanthology.org/2024.naacl-long.126/",
    doi = "10.18653/v1/2024.naacl-long.126",
    pages = "2279--2292"
}

@article{truth-is-lie,
title = "Truth Is a Lie: Crowd Truth and the Seven Myths of Human Annotation",
author = "Lora Aroyo and Chris Welty",
year = "2015",
month = mar,
day = "1",
doi = "10.1609/aimag.v36i1.2564",
language = "English",
volume = "36",
pages = "15--24",
journal = "The AI Magazine",
issn = "0738-4602",
publisher = "John Wiley and Sons Inc.",
number = "1",
}

@misc{stance1,
      title={Whose Opinions Do Language Models Reflect?}, 
      author={Shibani Santurkar and Esin Durmus and Faisal Ladhak and Cinoo Lee and Percy Liang and Tatsunori Hashimoto},
      year={2023},
      eprint={2303.17548},
      archivePrefix={arXiv},
      primaryClass={cs.CL},
      url={https://arxiv.org/abs/2303.17548}, 
}

@misc{stance2,
      title={Beyond prompt brittleness: Evaluating the reliability and consistency of political worldviews in LLMs}, 
      author={Tanise Ceron and Neele Falk and Ana Barić and Dmitry Nikolaev and Sebastian Padó},
      year={2024},
      eprint={2402.17649},
      archivePrefix={arXiv},
      primaryClass={cs.CL},
      url={https://arxiv.org/abs/2402.17649}, 
}

@misc{stance3,
      title={"We Demand Justice!": Towards Social Context Grounding of Political Texts}, 
      author={Rajkumar Pujari and Chengfei Wu and Dan Goldwasser},
      year={2024},
      eprint={2311.09106},
      archivePrefix={arXiv},
      primaryClass={cs.CL},
      url={https://arxiv.org/abs/2311.09106}, 
}

@article{persp-survey,
author = {Frenda, Simona and Abercrombie, Gavin and Basile, Valerio and Pedrani, Alessandro and Panizzon, Raffaella and Cignarella, Alessandra Teresa and Marco, Cristina and Bernardi, Davide},
title = {Perspectivist approaches to natural language processing: a survey: Perspectivist approaches to natural language processing...},
year = {2024},
issue_date = {Jun 2025},
publisher = {Springer-Verlag},
address = {Berlin, Heidelberg},
volume = {59},
number = {2},
issn = {1574-020X},
url = {https://doi.org/10.1007/s10579-024-09766-4},
doi = {10.1007/s10579-024-09766-4},
journal = {Lang. Resour. Eval.},
month = aug,
pages = {1719–1746},
numpages = {28}
}

@inproceedings{convokit,
  title={ConvoKit: A Toolkit for the Analysis of Conversations},
  author={Chang, Jonathan P. and Chiam, Caleb and Fu, Liye and Wang, Andrew and Zhang, Justine and Danescu-Niculescu-Mizil, Cristian},
  booktitle={Proceedings of SIGDIAL},
  year={2020},
}

@misc{nadeem2020stereosetmeasuringstereotypicalbias,
      title={StereoSet: Measuring stereotypical bias in pretrained language models}, 
      author={Moin Nadeem and Anna Bethke and Siva Reddy},
      year={2020},
      eprint={2004.09456},
      archivePrefix={arXiv},
      primaryClass={cs.CL},
      url={https://arxiv.org/abs/2004.09456}, 
}

@misc{sap2020socialbiasframesreasoning,
      title={Social Bias Frames: Reasoning about Social and Power Implications of Language}, 
      author={Maarten Sap and Saadia Gabriel and Lianhui Qin and Dan Jurafsky and Noah A. Smith and Yejin Choi},
      year={2020},
      eprint={1911.03891},
      archivePrefix={arXiv},
      primaryClass={cs.CL},
      url={https://arxiv.org/abs/1911.03891}, 
}

@inproceedings{wood-doughty-etal-2021-using,
    title = "Using Noisy Self-Reports to Predict {T}witter User Demographics",
    author = "Wood-Doughty, Zach  and
      Xu, Paiheng  and
      Liu, Xiao  and
      Dredze, Mark",
    editor = "Ku, Lun-Wei  and
      Li, Cheng-Te",
    booktitle = "Proceedings of the Ninth International Workshop on Natural Language Processing for Social Media",
    month = jun,
    year = "2021",
    address = "Online",
    publisher = "Association for Computational Linguistics",
    url = "https://aclanthology.org/2021.socialnlp-1.11/",
    doi = "10.18653/v1/2021.socialnlp-1.11",
    pages = "123--137"
}

@inproceedings{sachdeva-etal-2022-targeted,
    title = "Targeted Identity Group Prediction in Hate Speech Corpora",
    author = "Sachdeva, Pratik  and
      Barreto, Renata  and
      Von Vacano, Claudia  and
      Kennedy, Chris",
    editor = "Narang, Kanika  and
      Mostafazadeh Davani, Aida  and
      Mathias, Lambert  and
      Vidgen, Bertie  and
      Talat, Zeerak",
    booktitle = "Proceedings of the Sixth Workshop on Online Abuse and Harms (WOAH)",
    month = jul,
    year = "2022",
    address = "Seattle, Washington (Hybrid)",
    publisher = "Association for Computational Linguistics",
    url = "https://aclanthology.org/2022.woah-1.22/",
    doi = "10.18653/v1/2022.woah-1.22",
    pages = "231--244"
}

@inproceedings{preotiuc-pietro-ungar-2018-user,
    title = "User-Level Race and Ethnicity Predictors from {T}witter Text",
    author = "Preo{\c{t}}iuc-Pietro, Daniel  and
      Ungar, Lyle",
    editor = "Bender, Emily M.  and
      Derczynski, Leon  and
      Isabelle, Pierre",
    booktitle = "Proceedings of the 27th International Conference on Computational Linguistics",
    month = aug,
    year = "2018",
    address = "Santa Fe, New Mexico, USA",
    publisher = "Association for Computational Linguistics",
    url = "https://aclanthology.org/C18-1130/",
    pages = "1534--1545"
}

@article{chesley_you_2011,
	title = {You {Know} {What} {It} {Is}: {Learning} {Words} through {Listening} to {Hip}-{Hop}},
	volume = {6},
	issn = {1932-6203},
	shorttitle = {You {Know} {What} {It} {Is}},
	url = {https://www.ncbi.nlm.nih.gov/pmc/articles/PMC3244393/},
	doi = {10.1371/journal.pone.0028248},
	number = {12},
	urldate = {2025-07-07},
	journal = {PLoS ONE},
	author = {Chesley, Paula},
	month = dec,
	year = {2011},
	pmid = {22205942},
	pmcid = {PMC3244393},
	pages = {e28248},
}

@inproceedings{ishibashi_subspace_2024,
	address = {Mexico City, Mexico},
	title = {Subspace {Representations} for {Soft} {Set} {Operations} and {Sentence} {Similarities}},
	url = {https://aclanthology.org/2024.naacl-long.192/},
	doi = {10.18653/v1/2024.naacl-long.192},
	urldate = {2025-07-07},
	booktitle = {Proceedings of the 2024 {Conference} of the {North} {American} {Chapter} of the {Association} for {Computational} {Linguistics}: {Human} {Language} {Technologies} ({Volume} 1: {Long} {Papers})},
	publisher = {Association for Computational Linguistics},
	author = {Ishibashi, Yoichi and Yokoi, Sho and Sudoh, Katsuhito and Nakamura, Satoshi},
	editor = {Duh, Kevin and Gomez, Helena and Bethard, Steven},
	month = jun,
	year = {2024},
	pages = {3512--3524},
}

@inproceedings{stewart_now_2014,
	address = {Gothenburg, Sweden},
	title = {Now {We} {Stronger} than {Ever}: {African}-{American} {English} {Syntax} in {Twitter}},
	shorttitle = {Now {We} {Stronger} than {Ever}},
	url = {https://aclanthology.org/E14-3004/},
	doi = {10.3115/v1/E14-3004},
	urldate = {2025-07-07},
	booktitle = {Proceedings of the {Student} {Research} {Workshop} at the 14th {Conference} of the {European} {Chapter} of the {Association} for {Computational} {Linguistics}},
	publisher = {Association for Computational Linguistics},
	author = {Stewart, Ian},
	editor = {Wintner, Shuly and Elliott, Desmond and Garoufi, Konstantina and Kiela, Douwe and Vulić, Ivan},
	month = apr,
	year = {2014},
	pages = {31--37},
}

@incollection{maclagan_regional_2005,
	title = {Regional and {Social} {Variation}},
	isbn = {978-0-470-75485-6},
	url = {https://onlinelibrary.wiley.com/doi/abs/10.1002/9780470754856.ch2},
	language = {en},
	urldate = {2025-07-07},
	booktitle = {Clinical {Sociolinguistics}},
	publisher = {John Wiley \& Sons, Ltd},
	author = {Maclagan, Margaret},
	year = {2005},
	doi = {10.1002/9780470754856.ch2},
	note = {Section: 2},
	pages = {15--25},
}

@article{suyudi_grammatical_2023,
	title = {Grammatical {Analysis} of {African} {American} {Vernacular} {English} in {The} {Eminem} {Show} {Album}: {A} {Linguistics} {Perspective}},
	copyright = {http://creativecommons.org/licenses/by-sa/4.0},
	issn = {2549-9017, 2460-2280},
	shorttitle = {Grammatical {Analysis} of {African} {American} {Vernacular} {English} in {The} {Eminem} {Show} {Album}},
	url = {https://ejournal.iainkendari.ac.id/index.php/langkawi/article/view/5897},
	doi = {10.31332/lkw.v0i0.5897},
	language = {en},
	urldate = {2025-07-07},
	journal = {Langkawi: Journal of The Association for Arabic and English},
	author = {Suyudi, Ichwan and Wibowo, Agung Prasetyo and Pasha, Luthfi Chanafiah},
	month = jun,
	year = {2023},
	note = {Publisher: Institut Agama Islam Negeri Kendari},
	pages = {56},
}

@inproceedings{geng_inducing_2022,
	address = {Abu Dhabi, United Arab Emirates},
	title = {Inducing {Generalizable} and {Interpretable} {Lexica}},
	url = {https://aclanthology.org/2022.findings-emnlp.325/},
	doi = {10.18653/v1/2022.findings-emnlp.325},
	urldate = {2025-07-10},
	booktitle = {Findings of the {Association} for {Computational} {Linguistics}: {EMNLP} 2022},
	publisher = {Association for Computational Linguistics},
	author = {Geng, Yilin and Wu, Zetian and Santhosh, Roshan and Srivastava, Tejas and Ungar, Lyle and Sedoc, João},
	editor = {Goldberg, Yoav and Kozareva, Zornitsa and Zhang, Yue},
	month = dec,
	year = {2022},
	pages = {4430--4448},
}

@article{bucholtz_identity_2005,
	title = {Identity and interaction: a sociocultural linguistic approach},
	volume = {7},
	copyright = {https://journals.sagepub.com/page/policies/text-and-data-mining-license},
	issn = {1461-4456, 1461-7080},
	shorttitle = {Identity and interaction},
	url = {https://journals.sagepub.com/doi/10.1177/1461445605054407},
	doi = {10.1177/1461445605054407},
	language = {en},
	number = {4-5},
	urldate = {2025-07-10},
	journal = {Discourse Studies},
	author = {Bucholtz, Mary and Hall, Kira},
	month = oct,
	year = {2005},
	note = {Publisher: SAGE Publications},
	pages = {585--614},
}

@inproceedings{shoemark_inducing_2018,
	address = {Brussels, Belgium},
	title = {Inducing a lexicon of sociolinguistic variables from code-mixed text},
	url = {https://aclanthology.org/W18-6101/},
	doi = {10.18653/v1/W18-6101},
	urldate = {2025-06-09},
	booktitle = {Proceedings of the 2018 {EMNLP} {Workshop} {W}-{NUT}: {The} 4th {Workshop} on {Noisy} {User}-generated {Text}},
	publisher = {Association for Computational Linguistics},
	author = {Shoemark, Philippa and Kirby, James and Goldwater, Sharon},
	editor = {Xu, Wei and Ritter, Alan and Baldwin, Tim and Rahimi, Afshin},
	month = nov,
	year = {2018},
	pages = {1--6},
}

@inproceedings{ziems_value_2022,
	address = {Dublin, Ireland},
	title = {{VALUE}: {Understanding} {Dialect} {Disparity} in {NLU}},
	shorttitle = {{VALUE}},
	url = {https://aclanthology.org/2022.acl-long.258/},
	doi = {10.18653/v1/2022.acl-long.258},
	urldate = {2025-07-07},
	booktitle = {Proceedings of the 60th {Annual} {Meeting} of the {Association} for {Computational} {Linguistics} ({Volume} 1: {Long} {Papers})},
	publisher = {Association for Computational Linguistics},
	author = {Ziems, Caleb and Chen, Jiaao and Harris, Camille and Anderson, Jessica and Yang, Diyi},
	editor = {Muresan, Smaranda and Nakov, Preslav and Villavicencio, Aline},
	month = may,
	year = {2022},
	pages = {3701--3720},
}

@article{benor_talking_nodate,
	title = {Talking {Jewish}: {The} “{Ethnic} {English}” of {American} {Jews}},
	language = {en},
	author = {Benor, Sarah Bunin and Cohen, Steven M},
}

@article{tia_sociolinguistic_2020,
	title = {A {Sociolinguistic} {Investigation}},
	volume = {15},
	doi = {10.15294/lc.v15i1.25965},
	journal = {Language Circle Journal of Language and Literature},
	author = {Tia, Intan and Aryani, Intan},
	month = oct,
	year = {2020},
	pages = {67--72},
}

@book{wardhaugh_introduction_2021,
	series = {Blackwell {Textbooks} in {Linguistics}},
	title = {An {Introduction} to {Sociolinguistics}},
	isbn = {978-1-119-47354-1},
	url = {https://books.google.com/books?id=y0orEAAAQBAJ},
	publisher = {Wiley},
	author = {Wardhaugh, R. and Fuller, J.M.},
	year = {2021},
	lccn = {2021004385},
}

@inproceedings{hayati_does_2021,
    address = {Online and Punta Cana, Dominican Republic},
    title = {Does {BERT} {Learn} as {Humans} {Perceive}? {Understanding} {Linguistic} {Styles} through {Lexica}},
    shorttitle = {Does {BERT} {Learn} as {Humans} {Perceive}?},
    url = {https://aclanthology.org/2021.emnlp-main.510/},
    doi = {10.18653/v1/2021.emnlp-main.510},
    urldate = {2025-07-14},
    booktitle = {Proceedings of the 2021 {Conference} on {Empirical} {Methods} in {Natural} {Language} {Processing}},
    publisher = {Association for Computational Linguistics},
    author = {Hayati, Shirley Anugrah and Kang, Dongyeop and Ungar, Lyle},
    editor = {Moens, Marie-Francine and Huang, Xuanjing and Specia, Lucia and Yih, Scott Wen-tau},
    month = nov,
    year = {2021},
    pages = {6323--6331},
}

@inproceedings{pryzant_deconfounded_2018,
    address = {New Orleans, Louisiana},
    title = {Deconfounded {Lexicon} {Induction} for {Interpretable} {Social} {Science}},
    url = {https://aclanthology.org/N18-1146/},
    doi = {10.18653/v1/N18-1146},
    urldate = {2025-07-14},
    booktitle = {Proceedings of the 2018 {Conference} of the {North} {American} {Chapter} of the {Association} for {Computational} {Linguistics}: {Human} {Language} {Technologies}, {Volume} 1 ({Long} {Papers})},
    publisher = {Association for Computational Linguistics},
    author = {Pryzant, Reid and Shen, Kelly and Jurafsky, Dan and Wagner, Stefan},
    editor = {Walker, Marilyn and Ji, Heng and Stent, Amanda},
    month = jun,
    year = {2018},
    pages = {1615--1625},
}

@book{boyd2022development,
  author    = {Boyd, Rod L. and Ashokkumar, Aneesh and Seraj, Sara and Pennebaker, James W.},
  title     = {The Development and Psychometric Properties of {LIWC-22}},
  year      = {2022},
  publisher = {University of Texas at Austin},
  address   = {Austin, TX},
}

@misc{havaldar_building_2024,
    title = {Building {Knowledge}-{Guided} {Lexica} to {Model} {Cultural} {Variation}},
    url = {http://arxiv.org/abs/2406.11622},
    doi = {10.48550/arXiv.2406.11622},
    urldate = {2025-07-14},
    publisher = {arXiv},
    author = {Havaldar, Shreya and Giorgi, Salvatore and Rai, Sunny and Cho, Young-Min and Talhelm, Thomas and Guntuku, Sharath Chandra and Ungar, Lyle},
    month = oct,
    year = {2024},
    note = {arXiv:2406.11622 [cs]
version: 2},
}

@inproceedings{ribeiro_why_2016,
    address = {San Diego, California},
    title = {“{Why} {Should} {I} {Trust} {You}?”: {Explaining} the {Predictions} of {Any} {Classifier}},
    shorttitle = {“{Why} {Should} {I} {Trust} {You}?},
    url = {http://aclweb.org/anthology/N16-3020},
    doi = {10.18653/v1/n16-3020},
    urldate = {2025-07-14},
    booktitle = {Proceedings of the 2016 {Conference} of the {North} {American} {Chapter} of the {Association} for {Computational} {Linguistics}: {Demonstrations}},
    publisher = {Association for Computational Linguistics},
    author = {Ribeiro, Marco and Singh, Sameer and Guestrin, Carlos},
    year = {2016},
}

@inproceedings{kim_interpretation_2020,
    address = {Online},
    title = {Interpretation of {NLP} models through input marginalization},
    url = {https://www.aclweb.org/anthology/2020.emnlp-main.255},
    doi = {10.18653/v1/2020.emnlp-main.255},
    urldate = {2025-07-14},
    booktitle = {Proceedings of the 2020 {Conference} on {Empirical} {Methods} in {Natural} {Language} {Processing} ({EMNLP})},
    publisher = {Association for Computational Linguistics},
    author = {Kim, Siwon and Yi, Jihun and Kim, Eunji and Yoon, Sungroh},
    year = {2020},
}

@misc{lundberg_unified_2017,
    title = {A {Unified} {Approach} to {Interpreting} {Model} {Predictions}},
    url = {http://arxiv.org/abs/1705.07874},
    doi = {10.48550/arXiv.1705.07874},
    urldate = {2025-07-14},
    publisher = {arXiv},
    author = {Lundberg, Scott and Lee, Su-In},
    month = nov,
    year = {2017},
    note = {arXiv:1705.07874 [cs]},
}

@misc{khattab_colbert_2020,
    title = {{ColBERT}: {Efficient} and {Effective} {Passage} {Search} via {Contextualized} {Late} {Interaction} over {BERT}},
    shorttitle = {{ColBERT}},
    url = {http://arxiv.org/abs/2004.12832},
    doi = {10.48550/arXiv.2004.12832},
    urldate = {2025-07-14},
    publisher = {arXiv},
    author = {Khattab, Omar and Zaharia, Matei},
    month = jun,
    year = {2020},
    note = {arXiv:2004.12832 [cs]},
}

@inproceedings{jorgensen_challenges_2015,
    address = {Beijing, China},
    title = {Challenges of studying and processing dialects in social media},
    url = {https://aclanthology.org/W15-4302/},
    doi = {10.18653/v1/W15-4302},
    urldate = {2025-07-15},
    booktitle = {Proceedings of the {Workshop} on {Noisy} {User}-generated {Text}},
    publisher = {Association for Computational Linguistics},
    author = {Jørgensen, Anna and Hovy, Dirk and Søgaard, Anders},
    editor = {Xu, Wei and Han, Bo and Ritter, Alan},
    month = jul,
    year = {2015},
    pages = {9--18},
}

@book{mcwhorter_word_2009,
    title = {Word {On} {The} {Street}: {Debunking} {The} {Myth} {Of} {A} {Pure} {Standard} {English}},
    isbn = {978-0-7867-3147-3},
    url = {https://books.google.com/books?id=bvs4DgAAQBAJ},
    publisher = {Basic Books},
    author = {Mcwhorter, John},
    year = {2009},
}

@book{smitherman_african_2007,
    title = {African {American} {English}},
    isbn = {978-3-638-63147-1},
    url = {https://books.google.com/books?id=CC-HhBoVlpYC},
    publisher = {GRIN Verlag},
    author = {Smitherman, Geneva},
    year = {2007},
}

@book{benor_becoming_2012,
    series = {Jewish {Cultures} of the {World}},
    title = {Becoming {Frum}: {How} {Newcomers} {Learn} the {Language} and {Culture} of {Orthodox} {Judaism}},
    isbn = {978-0-8135-5391-7},
    url = {https://books.google.com/books?id=IhkztUVgVRUC},
    publisher = {Rutgers University Press},
    author = {Benor, S.B.},
    year = {2012},
    lccn = {2012000002},
}

@article{mcwhorter2013talking,
  author   = {McWhorter, John},
  title    = {Talking Like That},
  journal  = {Jewish Review of Books},
  year     = {2013},
  month    = {Summer},
  url      = {https://jewishreviewofbooks.com/articles/401/talking-like-that/},
  urldate  = {2025-07-19},
}

@article{banna_cross-cultural_2016,
    title = {Cross-cultural comparison of perspectives on healthy eating among {Chinese} and {American} undergraduate students},
    volume = {16},
    issn = {1471-2458},
    url = {https://doi.org/10.1186/s12889-016-3680-y},
    doi = {10.1186/s12889-016-3680-y},
    number = {1},
    urldate = {2025-07-21},
    journal = {BMC Public Health},
    author = {Banna, Jinan C. and Gilliland, Betsy and Keefe, Margaret and Zheng, Dongping},
    month = sep,
    year = {2016},
    pages = {1015},
}

@article{tia_african_2020,
    title = {African {American} {Vernacular} {English} ({AAVE}) {Used} by {Rich} {Brian}: {A} {Sociolinguistic} {Investigation}},
    volume = {15},
    doi = {10.15294/lc.v15i1.25965},
    journal = {Language Circle Journal of Language and Literature},
    author = {Tia, Intan and Aryani, Intan},
    month = oct,
    year = {2020},
    pages = {67--72},
}

@misc{li_studying_2020,
    title = {Studying {Politeness} across {Cultures} {Using} {English} {Twitter} and {Mandarin} {Weibo}},
    url = {http://arxiv.org/abs/2008.02449},
    doi = {10.48550/arXiv.2008.02449},
    urldate = {2025-07-21},
    publisher = {arXiv},
    author = {Li, Mingyang and Hickman, Louis and Tay, Louis and Ungar, Lyle and Guntuku, Sharath Chandra},
    month = aug,
    year = {2020},
    note = {arXiv:2008.02449 [cs]},
}

@misc{havaldar_comparing_2025,
    title = {Comparing {Styles} across {Languages}: {A} {Cross}-{Cultural} {Exploration} of {Politeness}},
    shorttitle = {Comparing {Styles} across {Languages}},
    url = {http://arxiv.org/abs/2310.07135},
    doi = {10.48550/arXiv.2310.07135},
    urldate = {2025-07-21},
    publisher = {arXiv},
    author = {Havaldar, Shreya and Pressimone, Matthew and Wong, Eric and Ungar, Lyle},
    month = mar,
    year = {2025},
    note = {arXiv:2310.07135 [cs]},
}

@inproceedings{roy_tale_2023,
    address = {Singapore},
    title = {“{A} {Tale} of {Two} {Movements}': {Identifying} and {Comparing} {Perspectives} in \#{BlackLivesMatter} and \#{BlueLivesMatter} {Movements}-related {Tweets} using {Weakly} {Supervised} {Graph}-based {Structured} {Prediction}},
    shorttitle = {“{A} {Tale} of {Two} {Movements}'},
    url = {https://aclanthology.org/2023.findings-emnlp.701/},
    doi = {10.18653/v1/2023.findings-emnlp.701},
    urldate = {2025-07-21},
    booktitle = {Findings of the {Association} for {Computational} {Linguistics}: {EMNLP} 2023},
    publisher = {Association for Computational Linguistics},
    author = {Roy, Shamik and Goldwasser, Dan},
    editor = {Bouamor, Houda and Pino, Juan and Bali, Kalika},
    month = dec,
    year = {2023},
    pages = {10437--10467},
}

@misc{borenstein_investigating_2024,
    title = {Investigating {Human} {Values} in {Online} {Communities}},
    url = {http://arxiv.org/abs/2402.14177},
    doi = {10.48550/arXiv.2402.14177},
    urldate = {2025-07-21},
    publisher = {arXiv},
    author = {Borenstein, Nadav and Arora, Arnav and Kaffee, Lucie-Aimée and Augenstein, Isabelle},
    month = nov,
    year = {2024},
    note = {arXiv:2402.14177 [cs]},
}

@inproceedings{reuver_topic-specific_2024,
    address = {Vienna, Austria},
    title = {Topic-specific social science theory in stance detection: a proposal and interdisciplinary pilot study on sustainability initiatives},
    shorttitle = {Topic-specific social science theory in stance detection},
    url = {https://aclanthology.org/2024.cpss-1.8/},
    urldate = {2025-07-21},
    booktitle = {Proceedings of the 4th {Workshop} on {Computational} {Linguistics} for the {Political} and {Social} {Sciences}: {Long} and short papers},
    publisher = {Association for Computational Linguistics},
    author = {Reuver, Myrthe and Polimeno, Alessandra and Fokkens, Antske and Lopes, Ana Isabel},
    editor = {Klamm, Christopher and Lapesa, Gabriella and Ponzetto, Simone Paolo and Rehbein, Ines and Sen, Indira},
    month = sep,
    year = {2024},
    pages = {101--111},
}

@article{alliheibi_opinion_2021,
    title = {Opinion {Mining} of {Saudi} {Responses} to {COVID}-19 {Vaccines} on {Twitter}},
    volume = {12},
    issn = {2156-5570},
    url = {https://thesai.org/Publications/ViewPaper?Volume=12&Issue=6&Code=IJACSA&SerialNo=10},
    doi = {10.14569/IJACSA.2021.0120610},
    language = {en},
    number = {6},
    urldate = {2025-07-21},
    journal = {International Journal of Advanced Computer Science and Applications (IJACSA)},
    author = {Alliheibi, Fahad M. and Omar, Abdulfattah and Al-Horais, Nasser},
    year = {2021},
    note = {Number: 6
Publisher: The Science and Information (SAI) Organization Limited},
}

@INPROCEEDINGS{Lin_etal_SIGIR2021_Pyserini,
   author = "Jimmy Lin and Xueguang Ma and Sheng-Chieh Lin and Jheng-Hong Yang and Ronak Pradeep and Rodrigo Nogueira",
   title = "{Pyserini}: A {Python} Toolkit for Reproducible Information Retrieval Research with Sparse and Dense Representations",
   booktitle = "Proceedings of the 44th Annual International ACM SIGIR Conference on Research and Development in Information Retrieval (SIGIR 2021)",
   year = 2021,
   pages = "2356--2362",
}

@inproceedings{caplan_conceptcarve_2025,
    title = "{C}oncept{C}arve: Dynamic Realization of Evidence",
    author = "Caplan, Eylon  and
      Goldwasser, Dan",
    editor = "Che, Wanxiang  and
      Nabende, Joyce  and
      Shutova, Ekaterina  and
      Pilehvar, Mohammad Taher",
    booktitle = "Proceedings of the 63rd Annual Meeting of the Association for Computational Linguistics (Volume 1: Long Papers)",
    month = jul,
    year = "2025",
    address = "Vienna, Austria",
    publisher = "Association for Computational Linguistics",
    url = "https://aclanthology.org/2025.acl-long.1014/",
    doi = "10.18653/v1/2025.acl-long.1014",
    pages = "20792--20809",
    ISBN = "979-8-89176-251-0"
}

@inproceedings{astuti_use_2018,
    title = {{THE} {USE} {OF} {AFRICAN}-{AMERICAN} {VERNACULAR} {ENGLISH} ({AAVE}) {IN} {LOGIC}’{S} {EVERYBODY}},
    url = {https://www.semanticscholar.org/paper/THE-USE-OF-AFRICAN-AMERICAN-VERNACULAR-ENGLISH-IN-Astuti/d4041d258352d1bfbc33ec84d9db52fca1715c2c},
    urldate = {2025-07-24},
    author = {Astuti, Puput Puji},
    month = dec,
    year = {2018},
}

@book{tuffery_data_2011,
    series = {Wiley {Series} in {Computational} {Statistics}},
    title = {Data {Mining} and {Statistics} for {Decision} {Making}},
    isbn = {978-0-470-97928-0},
    url = {https://books.google.com/books?id=5MTBlxZUKiIC},
    publisher = {Wiley},
    author = {Tufféry, S.},
    year = {2011},
    lccn = {2010039789},
}

@misc{yang_large_2024,
    title = {Large {Language} {Models} for {Automated} {Open}-domain {Scientific} {Hypotheses} {Discovery}},
    url = {http://arxiv.org/abs/2309.02726},
    doi = {10.48550/arXiv.2309.02726},
    urldate = {2025-09-24},
    publisher = {arXiv},
    author = {Yang, Zonglin and Du, Xinya and Li, Junxian and Zheng, Jie and Poria, Soujanya and Cambria, Erik},
    month = jun,
    year = {2024},
    note = {arXiv:2309.02726 [cs]},
}

@misc{manning_automated_2024,
    title = {Automated {Social} {Science}: {Language} {Models} as {Scientist} and {Subjects}},
    shorttitle = {Automated {Social} {Science}},
    url = {http://arxiv.org/abs/2404.11794},
    doi = {10.48550/arXiv.2404.11794},
    urldate = {2025-09-24},
    publisher = {arXiv},
    author = {Manning, Benjamin S. and Zhu, Kehang and Horton, John J.},
    month = apr,
    year = {2024},
    note = {arXiv:2404.11794 [econ]},
}

@article{peterson_using_2021,
    title = {Using large-scale experiments and machine learning to discover theories of human decision-making},
    volume = {372},
    url = {https://www.science.org/doi/10.1126/science.abe2629},
    doi = {10.1126/science.abe2629},
    number = {6547},
    urldate = {2025-09-24},
    journal = {Science},
    author = {Peterson, Joshua C. and Bourgin, David D. and Agrawal, Mayank and Reichman, Daniel and Griffiths, Thomas L.},
    month = jun,
    year = {2021},
    note = {Publisher: American Association for the Advancement of Science},
    pages = {1209--1214},
}

@article{airhihenbuwa_cultural_1996,
    title = {Cultural aspects of {African} {American} eating patterns},
    url = {https://pubmed.ncbi.nlm.nih.gov/9395569/},
    language = {en},
    urldate = {2025-10-05},
    journal = {PubMed},
    author = {Airhihenbuwa, Co and Kumanyika, S and Agurs, Td and Lowe, A and Saunders, D and Morssink, Cb},
    year = {1996},
}

@article{cuevas_ag_african_2016,
    title = {African {American} experiences in healthcare: "{I} always feel like {I}'m getting skipped over"},
    shorttitle = {African {American} experiences in healthcare},
    url = {https://pubmed.ncbi.nlm.nih.gov/27175576/},
    language = {en},
    urldate = {2025-10-05},
    journal = {PubMed},
    author = {{Cuevas Ag} and {O'Brien K} and {Saha S}},
    year = {2016},
}

@incollection{fellezs_black_2012,
    title = {Black {Metal} {Soul} {Music}: {Stone} {Vengeance} and the {Aesthetics} of {Race} in {Heavy} {Metal}},
    isbn = {978-1-84888-147-1},
    shorttitle = {Black {Metal} {Soul} {Music}},
    author = {Fellezs, Kevin},
    month = apr,
    year = {2012},
    doi = {10.1163/9781848881471_004},
    pages = {23--36},
}

@article{caplan_public_2021,
    title = {Public {Heroes}, {Secret} {Jews}: {Jewish} {Identity} and {Comic} {Books}},
    volume = {14},
    issn = {1946-2522},
    shorttitle = {Public {Heroes}, {Secret} {Jews}},
    url = {https://muse.jhu.edu/pub/1/article/788411},
    number = {1},
    urldate = {2025-10-05},
    journal = {Journal of Jewish Identities},
    author = {Caplan, Jennifer},
    year = {2021},
    note = {Publisher: Johns Hopkins University Press},
    pages = {53--70},
}

@phdthesis{brunssen_making_2023,
    type = {Thesis},
    title = {The {Making} of "{Jew} {Clubs}": {Performing} {Jewishness} and {Antisemitism} in {European} {Soccer} and {Fan} {Cultures}},
    shorttitle = {The {Making} of "{Jew} {Clubs}"},
    url = {https://deepblue.lib.umich.edu/handle/2027.42/176679},
    language = {en\_US},
    urldate = {2025-10-05},
    author = {Brunssen, Pavel},
    year = {2023},
    doi = {10.7302/7528},
    note = {Accepted: 2023-05-25T14:56:53Z},
}

@article{stein_jewish_2019,
    title = {Jewish {Flow}: {Performing} {Identity} in {Hip}-{Hop} {Music}},
    volume = {38},
    issn = {0271-9274},
    shorttitle = {Jewish {Flow}},
    url = {https://www.jstor.org/stable/10.5325/studamerjewilite.38.2.0119},
    doi = {10.5325/studamerjewilite.38.2.0119},
    number = {2},
    urldate = {2025-10-05},
    journal = {Studies in American Jewish Literature (1981-)},
    author = {Stein, Leonard},
    year = {2019},
    note = {Publisher: Penn State University Press},
    pages = {119--139},
}

@article{campbell_gaming_2016,
    title = {Gaming {Religionworlds}: {Why} {Religious} {Studies} {Should} {Pay} {Attention} to {Religion} in {Gaming}},
    volume = {84},
    issn = {0002-7189},
    shorttitle = {Gaming {Religionworlds}},
    url = {https://www.jstor.org/stable/26177713},
    number = {3},
    urldate = {2025-10-05},
    journal = {Journal of the American Academy of Religion},
    author = {Campbell, Heidi A. and Wagner, Rachel and Luft, Shanny and Gregory, Rabia and Grieve, Gregory Price and Zeiler, Xenia},
    year = {2016},
    note = {Publisher: [Oxford University Press, American Academy of Religion]},
    pages = {641--664},
}

@article{czerwonka_application_2024,
    title = {Application of {Catholic} {Social} {Teaching} in {Finance} and {ManagementApplication} of {Catholic} {Social} {Teaching} in {Finance} and {Management}},
    volume = {14},
    doi = {10.15633/pch.14118},
    journal = {The Person and the Challenges. The Journal of Theology, Education, Canon Law and Social Studies Inspired by Pope John Paul II},
    author = {Czerwonka, Monika and Pietrzak, Maria},
    month = mar,
    year = {2024},
    pages = {295--313},
}

@article{jain_reinterpretation_2016,
    title = {Reinterpretation of {Hindu} {Myths} in {Contemporary} {Indian} {English} {Literature}},
    volume = {6},
    copyright = {Copyright (c) 2016 International Journal of Engineering, Science and Humanities},
    issn = {2250-3552},
    url = {https://www.ijesh.com/j/article/view/207},
    language = {en},
    number = {2},
    urldate = {2025-10-05},
    journal = {International Journal of Engineering, Science and Humanities},
    author = {Jain, Mahima},
    month = apr,
    year = {2016},
    pages = {08--14},
}

@phdthesis{cheung_learning_2015,
    type = {thesis},
    title = {Learning past the pictures in the panels: teacher attitudes to manga and anime texts},
    shorttitle = {Learning past the pictures in the panels},
    url = {https://figshare.mq.edu.au/articles/thesis/Learning_past_the_pictures_in_the_panels_teacher_attitudes_to_manga_and_anime_texts/19431809/1},
    language = {en},
    urldate = {2025-10-05},
    school = {Macquarie University},
    author = {Cheung, Kelly},
    year = {2015},
    doi = {10.25949/19431809.v1},
}

@article{hu_examination_2010,
    title = {An examination of blue- versus white-collar workers’ conceptualizations of job satisfaction facets},
    volume = {76},
    issn = {0001-8791},
    url = {https://www.sciencedirect.com/science/article/pii/S0001879109001419},
    doi = {10.1016/j.jvb.2009.10.014},
    number = {2},
    urldate = {2025-10-05},
    journal = {Journal of Vocational Behavior},
    author = {Hu, Xiaoxiao and Kaplan, Seth and Dalal, Reeshad S.},
    month = apr,
    year = {2010},
    pages = {317--325},
}

@article{mcdermott_thumbnail_2015,
    title = {A {Thumbnail} {Sketch} of {Judaism} for {Christians}},
    url = {https://www.cslewisinstitute.org/resources/a-thumbnail-sketch-of-judaism-for-christians/},
    language = {en-US},
    urldate = {2025-10-05},
    journal = {C.S. Lewis Institute},
    author = {McDermott, Gerald},
    year = {2015},
}

@article{schwartz_is_2011,
    title = {Is {There} {Life} {After} {Death}? {Jewish} {Thinking} on the {Afterlife}},
    shorttitle = {Is {There} {Life} {After} {Death}?},
    url = {https://momentmag.com/is-there-life-after-death/},
    language = {en-US},
    urldate = {2025-10-05},
    journal = {Moment Magazine},
    author = {Schwartz, Amy E.},
    month = jul,
    year = {2011},
}

@article{eckardt_death_1972,
    title = {Death in the {Judaic} and {Christian} {Traditions}},
    volume = {39},
    issn = {0037-783X},
    url = {https://www.jstor.org/stable/40970106},
    number = {3},
    urldate = {2025-10-05},
    journal = {Social Research},
    author = {Eckardt, A. Roy},
    year = {1972},
    note = {Publisher: The New School},
    pages = {489--514},
}

@book{gumperz1971language,
  title     = {Language in Social Groups},
  author    = {Gumperz, John J.},
  year      = {1971},
  publisher = {Stanford University Press},
  address   = {Stanford, CA}
}

@incollection{chambers2002studying,
  author    = {Chambers, J. K.},
  title     = {Studying Language Variation: An Informal Epistemology},
  booktitle = {The Handbook of Language Variation},
  editor    = {Chambers, J. K. and Trudgill, Peter and Schilling-Estes, Natalie},
  year      = {2002},
  publisher = {Blackwell},
  address   = {Oxford}
}

@book{jaffe2009sociolinguistic,
  editor    = {Jaffe, Alexandra},
  title     = {Sociolinguistic Perspectives on Stance},
  year      = {2009},
  publisher = {Oxford University Press},
  address   = {Oxford}
}

@article{eckert2012three,
  author    = {Eckert, Penelope},
  title     = {Three Waves of Variation Study: The Emergence of Meaning in the Study of Sociolinguistic Variation},
  journal   = {Annual Review of Anthropology},
  year      = {2012},
  volume    = {41},
  pages     = {87--100}
}

@article{hernandez-campoy_research_2014,
    title = {Research methods in {Sociolinguistics}},
    url = {https://www.jbe-platform.com/content/journals/10.1075/aila.27.01her},
    doi = {10.1075/aila.27.01her},
    language = {en},
    urldate = {2025-10-06},
    author = {Hernández-Campoy, Juan Manuel},
    month = jan,
    year = {2014},
    note = {Publisher: John Benjamins},
}

@inproceedings{tigunova_reddust_2020,
    address = {Marseille, France},
    title = {{RedDust}: a {Large} {Reusable} {Dataset} of {Reddit} {User} {Traits}},
    isbn = {979-10-95546-34-4},
    shorttitle = {{RedDust}},
    url = {https://aclanthology.org/2020.lrec-1.751/},
    language = {eng},
    urldate = {2025-10-06},
    booktitle = {Proceedings of the {Twelfth} {Language} {Resources} and {Evaluation} {Conference}},
    publisher = {European Language Resources Association},
    author = {Tigunova, Anna and Mirza, Paramita and Yates, Andrew and Weikum, Gerhard},
    editor = {Calzolari, Nicoletta and Béchet, Frédéric and Blache, Philippe and Choukri, Khalid and Cieri, Christopher and Declerck, Thierry and Goggi, Sara and Isahara, Hitoshi and Maegaard, Bente and Mariani, Joseph and Mazo, Hélène and Moreno, Asuncion and Odijk, Jan and Piperidis, Stelios},
    month = may,
    year = {2020},
    pages = {6118--6126},
}

\appendix
\section{Dataset Details}

\subsection{Demographic Processing}
\label{app:dataset-details}
\begin{figure}[h]
    \centering
    \includegraphics[width=\linewidth]{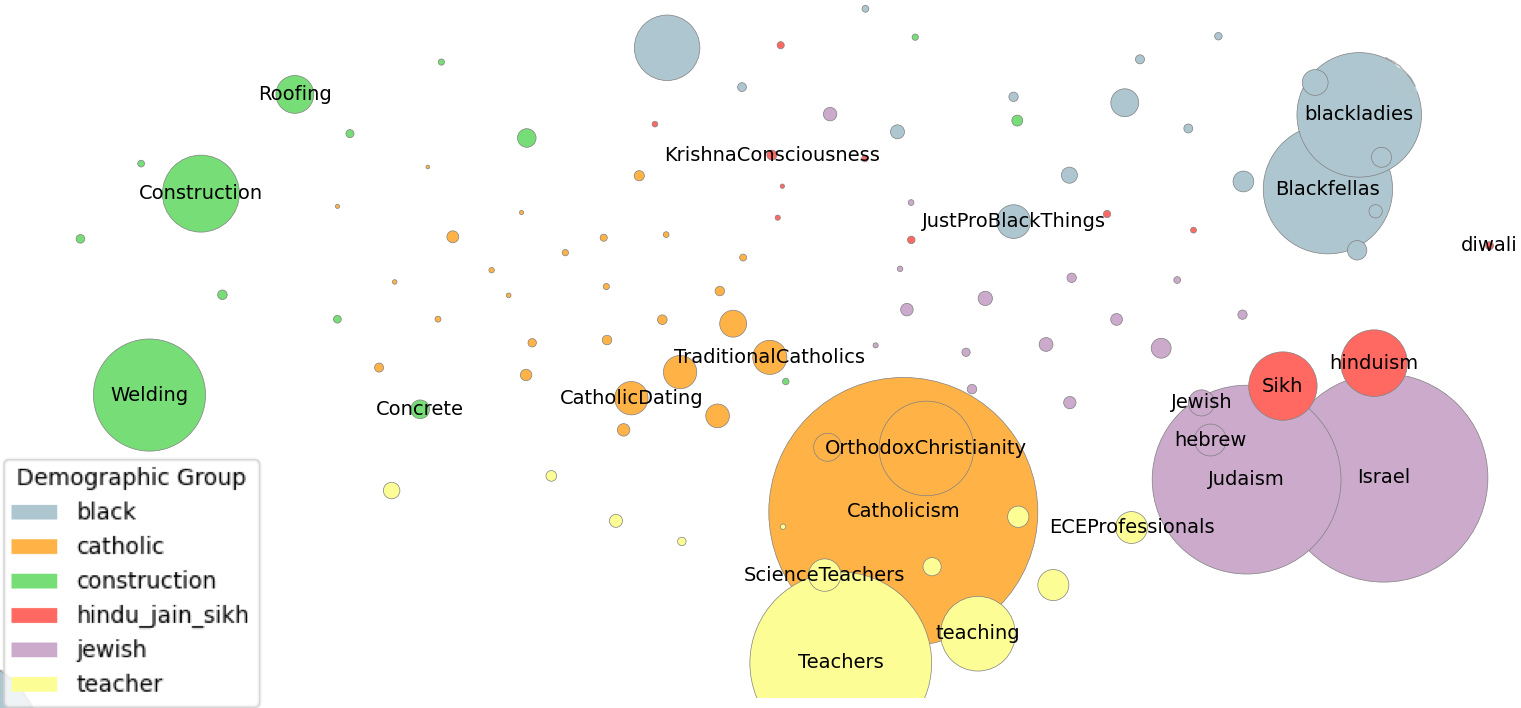}
    \caption{\small Visualization of the seed subreddit discovery process. Each bubble is a seed subreddit, sized by post volume and positioned by user overlap with other seeds, and clustering validates our iterative expansion method. Crucially, this plot shows raw user overlap between subreddits, not final demographic user groups, which are filtered to be nearly disjoint.}
    \label{fig:bubble-plot}
\end{figure}

\begin{table}[htbp]
\tiny
\centering
\begin{tabular}{ll}
\toprule
\textbf{Demographic} & \textbf{Seed subreddits} \\
\hline
African-American & 
\begin{tabular}[t]{@{}l@{}}
AfricanAmerican, BlackAtheism, \\
BlackHair, BlackLadiesFitness, \\
BlackWomens, Blackfellas, \\
\textbf{Blackpeople}, Blerds, \\
Dreadlocks, EbonyImagination, \\
JustProBlackThings, Natural\_Hair, \\
blackculture, blackgirlgamers, \\
blackgirls, blackinamerica, \\
blackladies, blackpower, \\
blackyoutubers
\end{tabular} \\
\hline
Catholics & 
\begin{tabular}[t]{@{}l@{}}
AnglicanOrdinariate, \textbf{Catholic}, \\
CatholicDating, CatholicGamers, \\
CatholicMemes, CatholicParenting, \\
CatholicPhilosophy, CatholicPolitics, \\
CatholicVideos, CatholicWomen, \\
Catholic\_News, Catholicism, \\
DeusVult, EasternCatholic, \\
FAMnNFP, MarriedCatholics, \\
OrthodoxChristianity, RCIA, \\
RealCatholicMen, Roman\_Catholics, \\
TraditionalCatholics, \\
TrueCatholicPolitics, VideoSancto, \\
catechism, divineoffice, \\
eRetreats, homilies, \\
knightsofcolumbus, \\
modelCatholicChurch
\end{tabular} \\
\hline
Teachers & 
\begin{tabular}[t]{@{}l@{}}
teachingresources, ELATeachers, \\
\textbf{Teachers}, historyteachers, \\
TeacherTales, SubstituteTeach, \\
teaching, Teacher, \\
ScienceTeachers, ArtEd, \\
ECEProfessionals, SpanishTeachers
\end{tabular} \\
\hline
Construction Workers & 
\begin{tabular}[t]{@{}l@{}}
Builders, Concrete, \\
Contractor, Insulation, \\
drywall, Welding, \\
Ironworker, \textbf{Construction}, \\
BlueCollarWomen, Roofing, \\
WeldPorn, weldingjobs, \\
ConstructionPorn, BadWelding
\end{tabular} \\
\hline
\makecell[lt]{Hindus \\ Jains \\ Sikhs} & 
\begin{tabular}[t]{@{}l@{}}
AdvaitaVedanta, \textbf{hindu}, \\
krishna, IndiaRWResources, \\
hinduism, KrishnaConsciousness, \\
hindurashtravad, Sikh, \\
Jainism, truehinduism, \\
diwali, bhajan
\end{tabular} \\
\bottomrule
\end{tabular}
\caption{Demographics and corresponding seed subreddits (first seed in bold)}
\label{tab:seed-subreddits}
\end{table}

\begin{table}[ht]
\tiny
\centering
\begin{tabular}{lll}
\toprule
\textbf{Demographic} & \textbf{Self-ID Phrases} & \textbf{Anti-Self-ID Phrases} \\ \hline
African-American & \begin{tabular}[t]{@{}l@{}}
I am black\\
I'm black\\
I am African American\\
...\\
As a black man\\
As a black woman
\end{tabular} & \begin{tabular}[t]{@{}l@{}}
I am not black\\
I'm not black\\
I'm white\\
...\\
I am Caucasian\\
I am Asian
\end{tabular} \\ \hline
Catholics & \begin{tabular}[t]{@{}l@{}}
I am Catholic\\
I'm Catholic\\
...\\
As a Roman Catholic\\
I belong to the Catholic Church\\
I am part of the Catholic Church
\end{tabular} & \begin{tabular}[t]{@{}l@{}}
I am not Catholic\\
I'm not Catholic\\
...\\
I am a Baptist\\
I'm Methodist\\
...\\
I am agnostic\\
As an ex-Christian\\
As an ex-Catholic
\end{tabular} \\ \hline
Teachers & \begin{tabular}[t]{@{}l@{}}
I am a teacher\\
I'm a teacher\\
I teach\\
I am an educator\\
...\\
I'm in education\\
I am a school teacher\\
I love teaching
\end{tabular} & \begin{tabular}[t]{@{}l@{}}
I am not a teacher\\
...\\
I am a student\\
...\\
I'm an accountant\\
I work in accounting\\
I am a scientist\\
...\\
I work in pharmacy
\end{tabular} \\ \hline
Construction \\Workers & \begin{tabular}[t]{@{}l@{}}
I am in construction\\
I'm in construction\\
...\\
I work in the construction industry\\
I'm in the construction industry\\
...\\
I work with drywall\\
I work in insulation\\
...\\
I am an ironworker\\
I'm an ironworker
\end{tabular} & \begin{tabular}[t]{@{}l@{}}
I am not in construction\\
I don't work construction\\
...\\
I have a desk job\\
I work in tech\\
I work in IT\\
I am a programmer\\
...\\
I am an artist\\
I'm an artist\\
I work in retail\\
I'm in retail
\end{tabular} \\ \hline
Hindus, Jains, \\ and Sikhs & \begin{tabular}[t]{@{}l@{}}
I am Hindu\\
I'm Hindu\\
As a Hindu\\
I identify as Hindu\\
...\\
I practice Sanatana Dharm\\
I follow Sanatana Dharm\\
I am a follower of Sanatana Dharm\\
...\\
I'm a follower of Sanatana Dharma\\
As a follower of Sanatana Dharma\\
I believe in Sanatana Dharma\\
I am Jain\\
I'm Jain\\
I am a Jain\\
I'm a Jain\\
...\\
I am Sikh\\
I'm Sikh\\
...\\
I practice Sikhism\\
As a follower of Sikhism
\end{tabular} & \begin{tabular}[t]{@{}l@{}}
I am not Hindu\\
I'm not Hindu\\
I am not a Hindu\\
I'm not a Hindu\\
...\\
I am Christian\\
I'm Christian\\
I am a Christian\\
...\\
I am agnostic\\
I have no religion\\
As an ex-Hindu\\
As an ex-Jain\\
As an ex-Sikh
\end{tabular} \\
\bottomrule
\end{tabular}
\caption{ Self-identification and anti self-identification phrases for various demographics}
\label{tab:demographics}
\end{table}

\begin{figure}[ht]
    \centering
    \includegraphics[width=\linewidth]{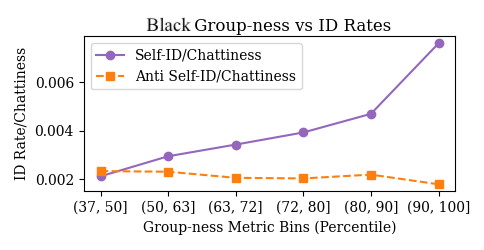}
    \label{fig:last_duck}
\end{figure}

\begin{figure*}[ht]
    \centering
    \begin{subfigure}[b]{0.49\linewidth}
        \centering
        \includegraphics[width=\linewidth]{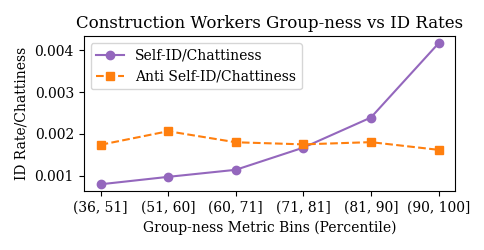}
    \end{subfigure}
    \hfill
    \begin{subfigure}[b]{0.49\linewidth}
        \centering
        \includegraphics[width=\linewidth]{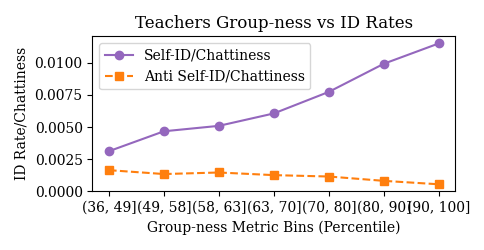}
    \end{subfigure}
    \hfill
    \begin{subfigure}[b]{0.49\linewidth}
        \centering
        \includegraphics[width=\linewidth]{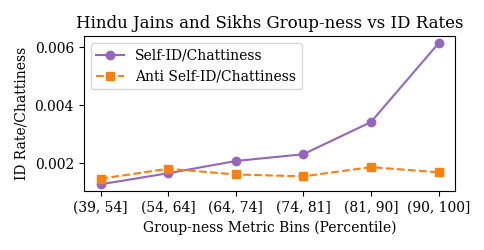}
    \end{subfigure}
    \hfill
    \begin{subfigure}[b]{0.49\linewidth}
        \centering
        \includegraphics[width=\linewidth]{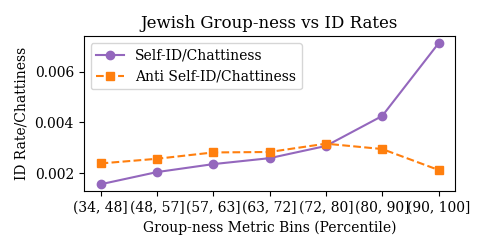}
    \end{subfigure}
    \caption{\small Normalized self-identification rate vs. group-ness of the remaining demographics.}
\label{fig:other_ducks}
\end{figure*}

\begin{figure}[h]
    \centering
    \begin{subfigure}[b]{\linewidth}
        \centering
        \includegraphics[width=\linewidth]{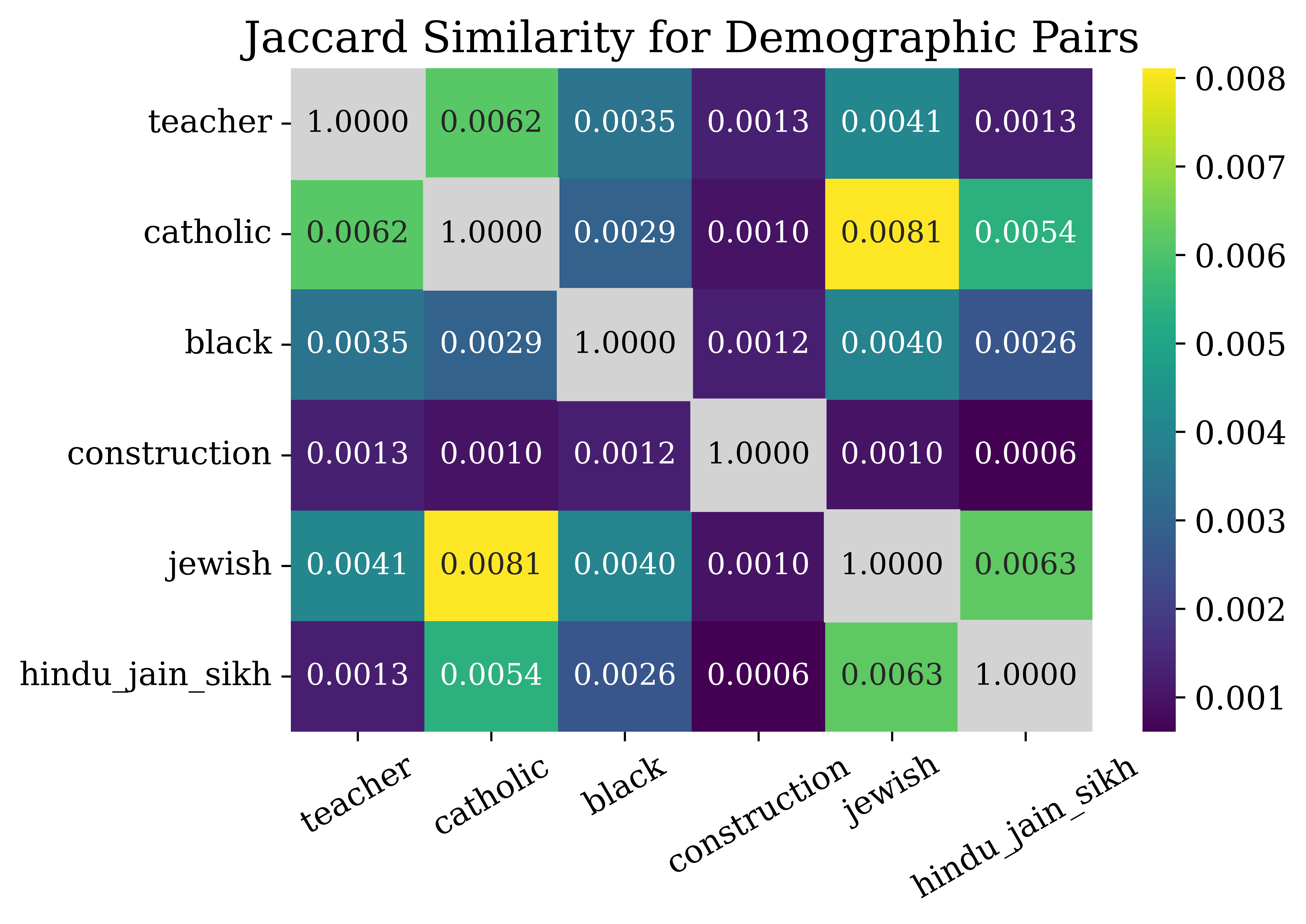}
    \end{subfigure}
    \caption{\small User intersectionality in the \splits dataset.}
    \label{fig:intersection-heatmap}
\end{figure}

Here we provide more details about the \splits dataset construction. Using both similarities (Jaccard/cosine) accounted for the skewed cases when one subreddit was very large while the other was very small. This is because the Jaccard index normalizes the intersection by the number of unique users in both subreddits, while the Cosine similarity normalizes by the geometric mean of the number of users. Table~\ref{tab:seed-subreddits} details the seed subreddits that were annotated for each demographic, and these subreddits are pictured in Figure~\ref{fig:bubble-plot}. Table~\ref{tab:demographics} details the self-identification and anti-self-identification phrases. We note that each $C_i$ was filtered to remove posts by known bots and then the top 1\% of users by number of posts were also filtered out to remove unknown bots and spam users. We also show the self-ID vs. anti-self-ID plots for the remaining demographics in Figure~\ref{fig:other_ducks}, and the exact thresholds used were: Jewish → 90, Black → 75, Catholic → 75, Construction → 90, Teacher → 75, Hindu/Jain/Sikh → 80, chosen based on separation in the plots. The full heatmap of demographic-demographic Jaccard similarities is in Figure~\ref{fig:intersection-heatmap}.

Two very common demographic dimensions, gender and age, were considered, but ultimately not used in \splits Because we aimed to limit intersectional identities, we opted not to use non-minority groups like `men' and `women' because they would result with huge amounts of overlap with other demographics. In addition, `women' subreddits were often labeled specifically for women (e.g. `r/TwoXChromosomes', `r/womenintech'), whereas they were very rarely for men, meaning selecting users in this way would almost certainly skew the distributions away from the real world.

For age, the single main issue was the demographic of people who use Reddit in general. Specifically, very few older-age subreddits exist, and the ones that exist have very few posts. Because of this, `age' as a demographic dimension was not used.

\subsection{When Demographic Annotation Fails: Koreans}
\label{app:korean-failure}
\begin{figure}[ht]
    \centering
    \includegraphics[width=\linewidth]{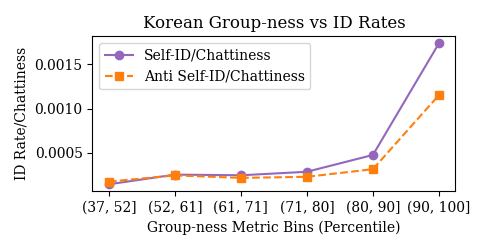}
    \caption{\small Failure case: normalized self-identification rate and anti self-identification rate do \textit{not} separate as group-ness increases.}
    \label{fig:korean_duck}
\end{figure}

Of the 7 demographic groups that we attempted to annotate, all were successful in having group-ness separate between self-ID and anti self-ID, \emph{except} for the group \textit{Koreans}. We annotated this set just as the others, starting with seed subreddit r/korea, and ending with seed set \{KLeague, KoreanFood, South\_Korea, busan, gyopo, hanguk, korea, seoul, southkorea\}. However, after collecting all posts across all Reddit, and computing the self-ID and anti self-ID rates, we saw a negative result (Figure~\ref{fig:korean_duck}). Here, we see that the anti self-ID rate does \textit{not} decrease as we turn up the group-ness, indicating that the group-ness metric here did not properly capture the target demographic. Inspecting the seed set, we hypothesize that this is because many non-Korean users were active in these groups such as Korean language learners, fans of Korean teams, or people interested in Korean culture like KoreanFood.

This demonstrates that our methodology using group-ness works only when the seed sets are `clean' enough so as to properly define a powerful group-ness metric for that demographic. Meaning, demographics unsuitable for our methodology are ones which do not have enough independent, clean sub-communities, making them difficult to isolate.

\subsection{Topic Processing}
\label{app:dataset-details-topic}

\begin{table*}[!ht]
\centering
\footnotesize
\begin{minipage}{0.5\textwidth}
\begin{center}
\begin{tabular}{l|l}
\hline
\textbf{Category} & \textbf{Topics} \\
\hline
Sports & basketball, soccer, football\ldots \\
Entertainment & superheroes, sci-fi, fantasy\ldots \\
Tech/Gaming & pc builds, coding, AI\ldots \\
Careers & jobs, resumes, freelance\ldots \\
Hobbies & gardening, cooking, crafts\ldots \\
Finance & budgets, stocks, retiring\ldots \\
Education & college, study tips, exams\ldots \\
News & global, politics, environment\ldots \\
Travel & budget, luxury, backpacking\ldots \\
Humor & memes, satire, animals\ldots \\
Political/Social & abortion, Russia, taxes\ldots \\
\hline
\end{tabular}
\caption{Categories and (some) topics.}
\label{tab:topics}
\end{center}
\end{minipage}%
\hfill
\begin{minipage}{0.46\textwidth}
\begin{center}
\begin{tabular}{l c c}
\hline
\textbf{Demographic} & \textbf{\# Posts (M)} & \textbf{\# Users (K)} \\
\hline
Teacher & 21.6 & 19.1 \\
Catholic & 21.5 & 13.5 \\
Black & 16.6 & 9.0 \\
\makecell[l]{Construction\\\hspace{1em}Worker} & 10.7 & 4.0 \\
Jewish & 15.2 & 5.3 \\
\makecell[l]{Hindu/Jain/\\\hspace{1em}Sikh} & 3.8 & 2.6 \\
\hline
\end{tabular}
\caption{Pre-topic filtration data stats.}
\label{tab:demographic_counts}
\end{center}
\end{minipage}
\end{table*}

Table~\ref{tab:topics} lists some categories and topics, while Table~\ref{tab:neutral_keywords} lists some select examples of topics within their categories, and shows the respective keywords used for the ColBERT topic relevance search. 

To cheaply ensure any given split's posts were truly about the topic in question, we applied the following process for each topic: first, we retrieved the top 100 thousand posts within each demographic's $C_i$ using ColBERT. We then combined all demographics' posts on the topic and ordered them by the ColBERT relevance score (600 thousand posts). Then we applied the LLM sliding window-binary search algorithm. The idea is as follows: in general, the ColBERT relevance score tells us relative relevance of documents to the topic. For example, in general it can tell us that post $x$ is more relevant to topic $t$ than post $y$. However, it does not at all tell us at what point posts are no longer relevant at all. Consider a contrived topic query like ``discussion of Christmas trees on the moon fighting over a purple golf club''. While we can assume there are either 0, or very few posts actually about this topic, the ColBERT retrieval will still give us 100 thousand posts for each demographic. So the point is to find the cutoff at which posts are (on average) no longer about the topic in question.

The binary search algorithm works as follows over a set of documents $D$ and topic $t$, given a window size $w$, a cutoff proportion $p$, and an error threshold $e$: the top $w$ posts are fed into an LLM one at a time, prompting it to say whether the post is within topic $t$ (yes/no). Among all posts in the window, the proportion of `yes's is computed. If the proportion is lower than $p$, then the window moves halfway upward between the current upper and lower bounds. Likewise, if the proportion is higher than $p$, then the window is moved halfway downward between the current lower and upper bound. This process continues until either a window's proportion is within $e$ of the desired cutoff $p$, or the bounds are exhausted. Once the process terminates, all posts ranked higher than the current window are considered above the cutoff, and included as `topical'.

For our processing, we used the gpt-4.1-nano-2025-04-14 model, $w=100, p=0.8,$ and $e=0.03$, meaning the bottom $100$ posts ranked by relevance of any split must have \emph{at least} 77\% topical posts as judged by the LLM. The underlying assumption for this algorithm is that \emph{on average}, higher ranked documents are more likely to be on topic that lower ranked documents. If it were exactly true, then we would not need a window. Due to the noisiness of the real-world data, we use a method more robust to this noise, while still ensuring that our LLM calls do not scale linearly with the number of posts (10's of millions).

\begin{figure}[h]
    \centering
    \begin{subfigure}[b]{\linewidth}
        \centering
        \includegraphics[width=\linewidth]{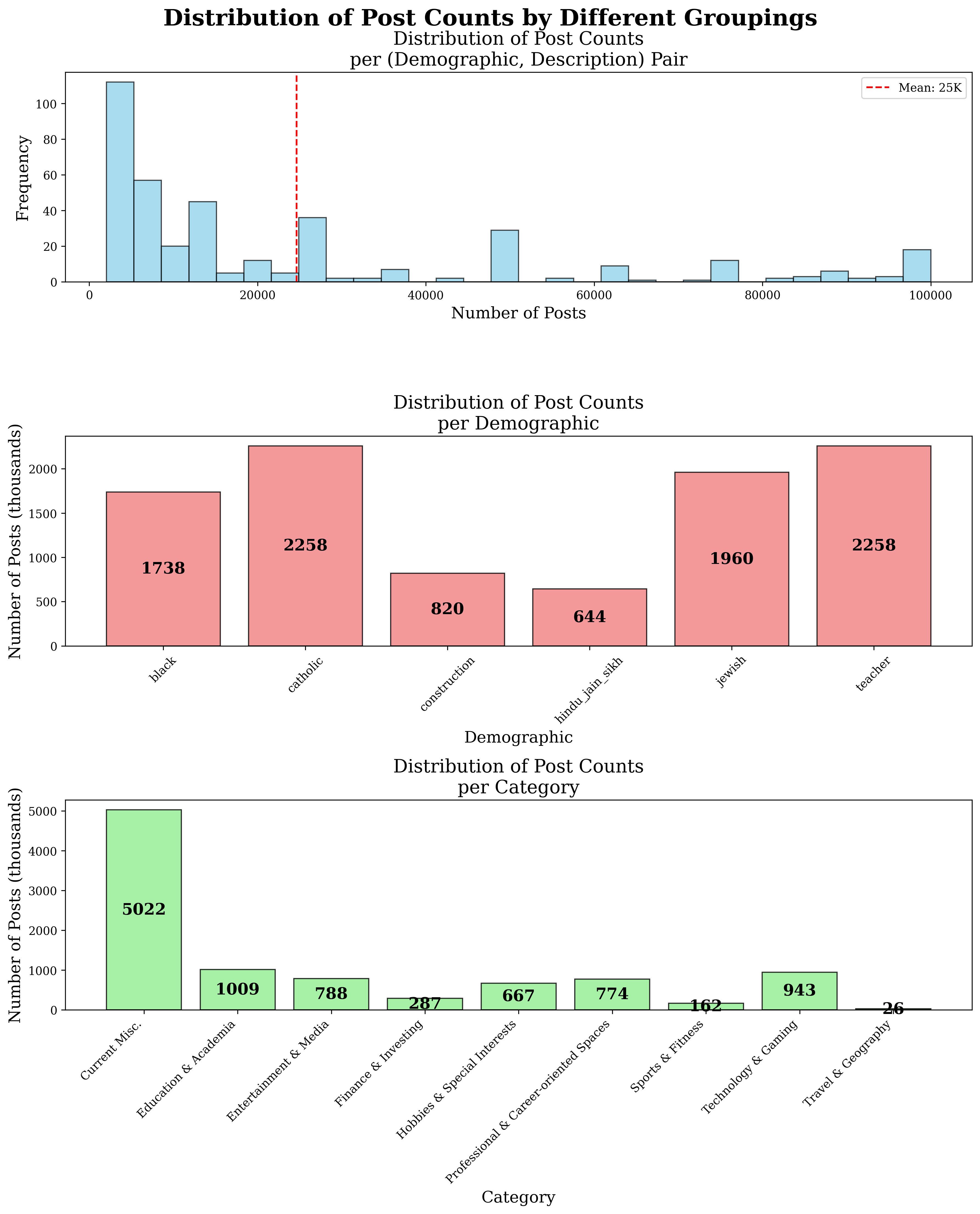}
    \end{subfigure}
    \caption{\small Distributions over different groupings of \splits.}
    \label{fig:post-count-distros}
\end{figure}

\begin{figure}[h]
    \centering
    \begin{subfigure}[b]{\linewidth}
        \centering
        \includegraphics[width=\linewidth]{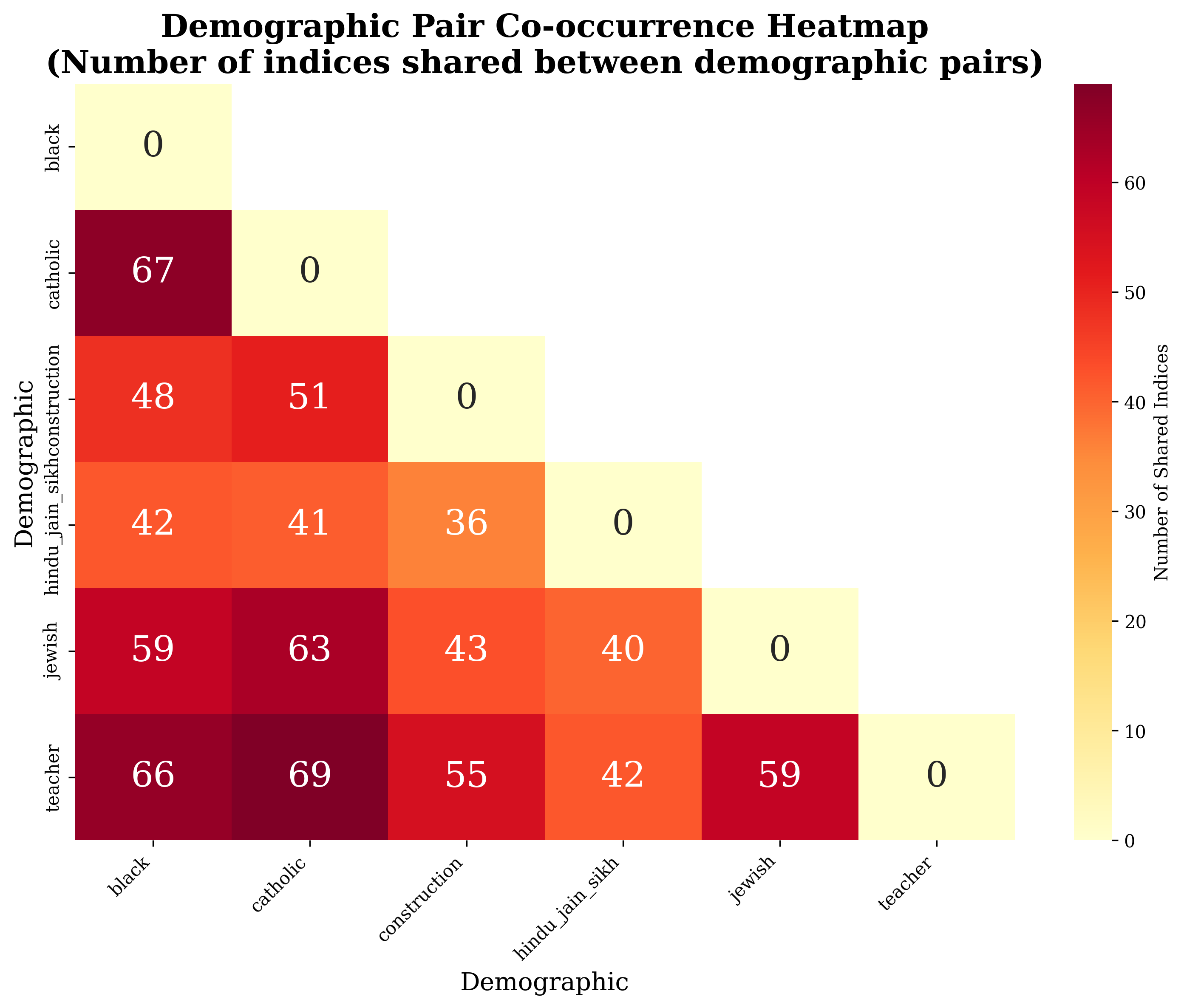}
    \end{subfigure}
    \caption{\small Heatmap of combined indices by demographic.}
    \label{fig:demo-index-heatmap}
\end{figure}

Figure~\ref{fig:post-count-distros} depicts the sizes of each $C_{d, t}$ split, both by demographic, topic category, and overall. The ``Political'' category is a catch all for many political and social topics of discussion, and includes more topics than the other categories (which is why it also has more posts). After removing splits with fewer than 2,000 posts, we also removed `orphan' splits that had only one demographic per topic. Figure~\ref{fig:demo-index-heatmap} shows the number of each pairing once the data is combined and indexed together. All pairs are very close in number. Figure~\ref{fig:triviality-vs-prompt} shows the exact distributions of triviality among all 5 kinds of prompts. This clearly shows how the more creative prompt leads to more non-trivial lexica, and that including information about the topic in the prompt helps reduce triviality.


\begin{table}[htbp]
\tiny
\centering
\begin{tabular}{lll}
\toprule
\textbf{Category} & \textbf{Specific Topic} & \textbf{Keywords} \\
\hline
\makecell[l]{Sports \\ \& Fitness} & Basketball & 
\begin{tabular}[t]{@{}l@{}}
basketball, hoop, net, dunk, \\ 
dribble, NBA\ldots
\end{tabular} \\
\hline
\makecell[l]{Sports \\ \& Fitness} & Soccer & 
\begin{tabular}[t]{@{}l@{}}
soccer, football, goal\ldots
\end{tabular} \\
\hline
\makecell[l]{Entertainment \\ \& Media} & \makecell[l]{Superheroes/Comic \\ Book Media} & 
\begin{tabular}[t]{@{}l@{}}
superheroes, comic books, \\ Marvel\ldots
\end{tabular} \\
\hline
\makecell[l]{Entertainment \\ \& Media} & Fantasy TV/Movies & 
\begin{tabular}[t]{@{}l@{}}
fantasy, magic, sword \\ and sorcery, 
medieval\ldots
\end{tabular} \\
\hline
\makecell[l]{Hobbies \\ \& Special Interests} & Gardening & 
\begin{tabular}[t]{@{}l@{}}
gardening, garden tips, \\plant care\ldots
\end{tabular} \\
\hline
\makecell[l]{Hobbies \\ \& Special Interests} & Cooking/Baking & 
\begin{tabular}[t]{@{}l@{}}
cooking, baking, recipes, \\ 
food blog\ldots
\end{tabular} \\
\hline
\makecell[l]{Education \\ \& Academia} & \makecell[l]{College Applications \\ \& Admissions} & 
\begin{tabular}[t]{@{}l@{}}
college applications, university \\ admissions, application process, \\
admission requirements, \\ 
college essay\ldots
\end{tabular} \\
\hline
\makecell[l]{Education \\ \& Academia} & \makecell[l]{Study Techniques \\ \& Productivity} & 
\begin{tabular}[t]{@{}l@{}}
study techniques, productivity \\ tips,
time management, study \\ schedule, 
note taking, active recall\ldots
\end{tabular} \\
\hline
\makecell[l]{Education \\ \& Academia} & \makecell[l]{Exam Preparation \\ \& Test-Taking Strategies} & 
\begin{tabular}[t]{@{}l@{}}
exam preparation, test strategies, \\ 
study tips, exam study guide, \\ 
test taking techniques\ldots
\end{tabular} \\
\bottomrule
\end{tabular}
\caption{Select example neutral categories, sub-topics and keywords}
\label{tab:neutral_keywords}
\end{table}

\subsection{Human Validation of Self-Identification Classification with LLM}
\label{app:human-val-self-id}
To compute self-identification and anti self-identification, we used first a set of regex expressions to retrieve posts, followed by an LLM filtration step to remove invalid matches (e.g. quotations, part of a bigger phrase, sarcasm, etc.). To assess the quality of this LLM filtration, we had an annotator manually examine 112 posts: 56 self-identification posts, and 56 anti self-identification posts, with each set having been labeled as ``valid matches'' by the LLM and the other half ``invalid''. The annotator also assessed the validity, and we report the results of using the \textit{regex only} vs. \textit{regex + LLM} in Table~\ref{tab:lift-vs-prompt}.

\begin{table}[h!]
\centering
\begin{tabular}{l ccc}
\hline
\textbf{Method} & \textbf{Precision} & \textbf{Recall} & \textbf{F1-Score} \\
\hline
Regex & 0.536 & 1.000 & 0.698 \\
Regex + LLM & 0.964 & 0.898 & 0.930 \\
\hline
\end{tabular}
\caption{Overall performance comparison of human annotation for self-ID/anti self-ID match validation. Adding the LLM greatly improves the precision, and therefore the F1 for less noisy results.}
\label{tab:regex-vs-regex-llm}
\end{table}

\subsection{Human Validation of Topic Classification with LLM}
\label{app:human-val-topic}
To ensure the quality of the topic annotation process, we randomly sampled 100 posts from diverse topics and demographics such that 50 were excluded as `not topical' and 50 were included as `topical'. We had a human annotator classify each post, and we compute the agreement with the LLM as a classifier. The result is [Precision: 0.96, Recall: 0.96, F1: 0.96]. Therefore we can say that the vast majority of topical annotations by the LLM are likely correct.
\section{Case Studies of Known SLPs}
\label{app:case-studies}

\textbf{AAVE} African American Vernacular English (AAVE) is a well-known and studied SLP \cite{stewart_now_2014, shoemark_inducing_2018, jorgensen_challenges_2015, ziems_value_2022}, which often involves \emph{code-switching}—using the vernacular more in some settings than others \cite{mcwhorter_word_2009}. For instance, AAVE is prevalent within the domain of Hip-Hop and Rap music \cite{tia_sociolinguistic_2020, suyudi_grammatical_2023, astuti_use_2018, chesley_you_2011}, while being used much less in professional settings. We test if our dataset captures these two patterns using a lexicon with features from \citet{ziems_value_2022} and \citet{smitherman_african_2007} (e.g., immediate future `finna', `ass' camouflage, copula deletion). We find that Black users use AAVE significantly more than non-Black users across both `Hip-Hop' and `Professional' topics. Furthermore, Black users themselves use AAVE features significantly more when discussing Hip-Hop than in professional contexts. Together, these results ($p<10^{-5}$) confirm that \splits is rich enough to capture both between-group linguistic phenomena and within-group code-switching.

\textbf{Jewish English} \citet{benor_talking_nodate} studied the vocabulary of American Jews, noting a difference in the usage of certain Yiddish and Hebrew words. Further, \citet{benor_becoming_2012, mcwhorter2013talking} study how Jews tend to use such in-group language less often when discussing secular topics as compared to more religious ones. Just as with Black AAVE use, we use the lexicon from \citet{benor_talking_nodate} to distinguish Jews from non-Jews, both on the topic of `Judaism', and on `Professional' topics. This again served the purpose of (1) verifying that the richness of the dataset captures the SLP of Jewish use of Yiddish words and (2) that this usage is code-switched depending on context.

We used the features from \citet{benor_talking_nodate}, including only lexicon entries that were used more by Jews than non-Jews. This included Yiddish/borrowed Hebrew terms. Fixing topic $t$ first as `Judaism', and then as `Professional' topics, for each demographic $d$ we compute the proportion of posts in $C_{d,t}$ with at least one Yiddish lexicon entry. As seen in Table~\ref{tab:yid_heb_mini}, we see that Jewish people in the dataset use Yiddish significantly more than non-Jewish people, both when discussing Judaism and also Professional topics, aligning with the SLP of Jewish people's Yiddish use. This answers question (1): the dataset \emph{is} sufficiently rich that Yiddish lexical features significantly distinguish between Jewish and non-Jewish users.

To answer question (2), we compared the proportions of only Jewish people's Yiddish use on the topic of Judaism vs. Professional topics. In that comparison, we also see a significantly higher use of Yiddish in Judaism than in Professional topics. This means that not only do Jewish people use Yiddish features more than non-Jewish users, but they use these features far more in certain contexts. These two results together show that the dataset captures the known SLP of (1) Jewish Yiddish use and (2) Jewish code-switching.

\begin{table}[ht]
  \scriptsize
  \centering
  \begin{tabular}{@{}llrr@{}}
    \toprule
    Group     & Topic   & Total posts & Prop.\ Yid/Heb \\
    \midrule
    J         & Jud.    & 96,880       & $1.90\times10^{-3}$ \\
    $\neg$J   & Jud.    & 64,400       & $7.30\times10^{-4}$ \\
    J         & Prof.   & 135,034      & $1.41\times10^{-4}$ \\
    $\neg$J   & Prof.   & 639,210      & $5.95\times10^{-5}$ \\
    \bottomrule
  \end{tabular}
  \vspace{1ex}

  \begin{tabular}{@{}lr@{}}
    \toprule
    Hypothesis                          & p-value    \\
    \midrule
    $p(J,Jud.)>p(\neg J,Jud.)$          & $<10^{-5}$ \\
    $p(J,Jud.)>p(J,Prof.)$              & $<10^{-5}$ \\
    $p(J,Prof.)>p(\neg J,Prof.)$        & $0.00079$  \\
    $p(\neg J,Jud.)>p(\neg J,Prof.)$    & $<10^{-5}$ \\
    \bottomrule
  \end{tabular}
  \caption{Yiddish/Hebrew usage stats and hypothesis tests}
  \label{tab:yid_heb_mini}
\end{table}

\begin{table}[ht]
  \scriptsize
  \centering
  \begin{tabular}{@{}llrr@{}}
    \toprule
    Group     & Topic & Total posts & Prop.\ dance \\
    \midrule
    HJS       & PCI   & 87,514       & 0.004365      \\
    $\neg$HJS & PCI   & 381,181      & 0.003618      \\
    \bottomrule
  \end{tabular}

  \vspace{1ex}

  \begin{tabular}{@{}lr@{}}
    \toprule
    Hypothesis                                   & p-value   \\
    \midrule
    $p(\mathrm{HJS},PCI)>p(\neg\mathrm{HJS},PCI)$ & $0.00054$ \\
    \bottomrule
  \end{tabular}
  \caption{``dance'' usage stats and hypothesis test}
  \label{tab:dance_mini}
\end{table}

\begin{figure}[h]
    \centering
    \begin{subfigure}[b]{\linewidth}
        \centering
        \includegraphics[width=\linewidth]{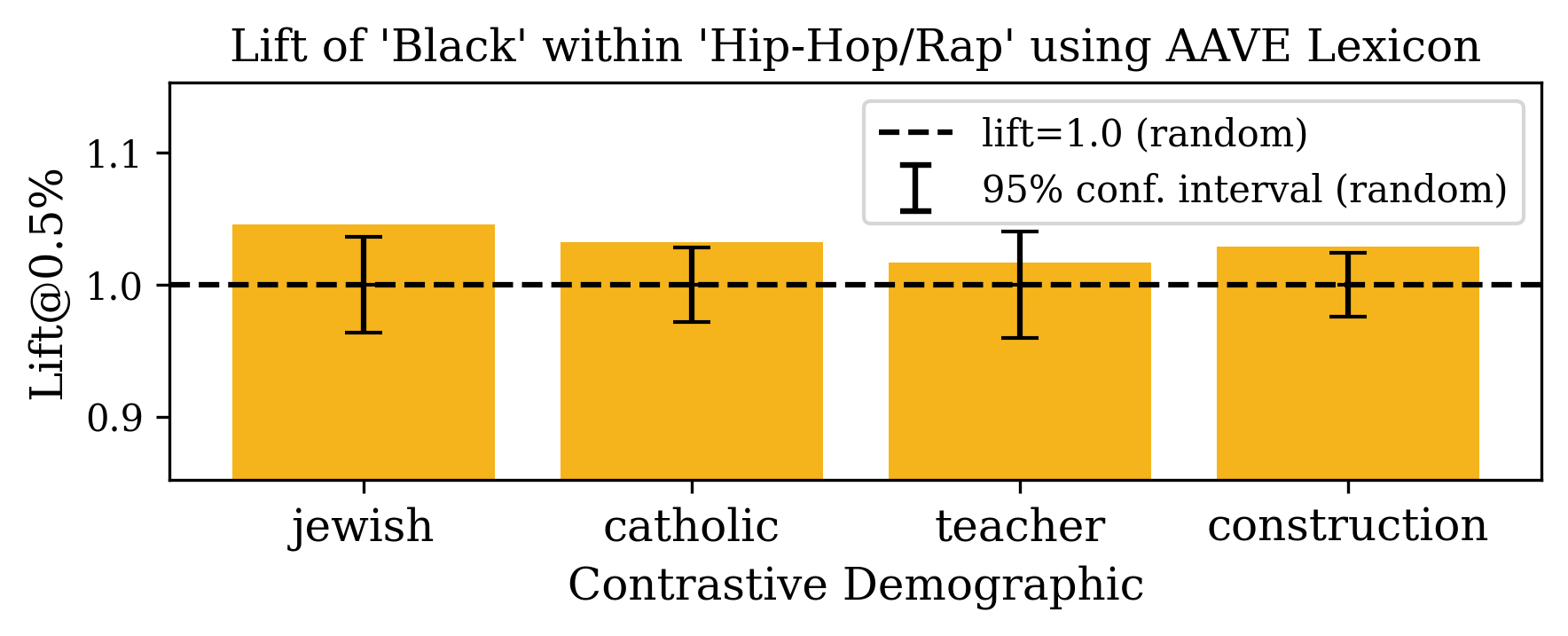}
    \end{subfigure}
    \caption{\small Lift at 0.5 of the Black demographic when talking about Hip-Hop/Rap using AAVE lexicon, as contrasted with 4 other demographics.}
    \label{fig:aave-lift}
\end{figure}

\begin{figure}[h]
    \centering
    \begin{subfigure}[b]{\linewidth}
        \centering
        \includegraphics[width=\linewidth]{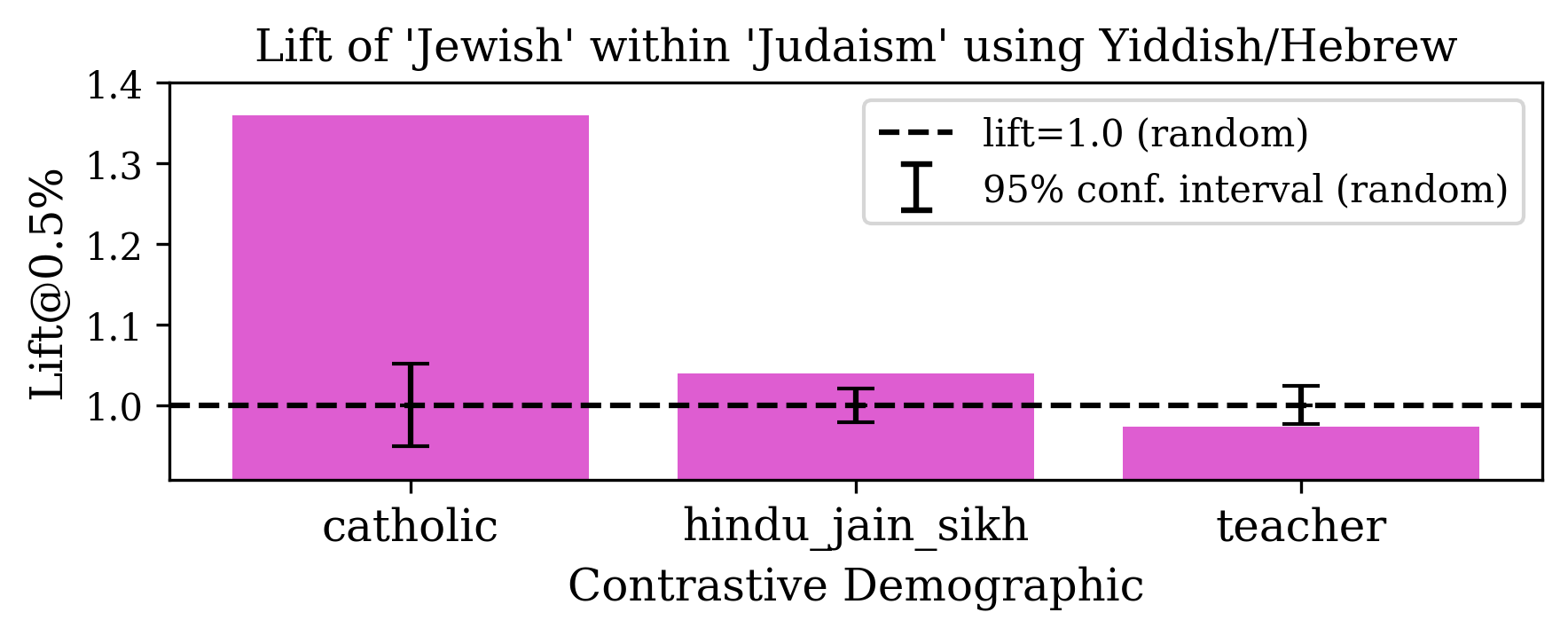}
    \end{subfigure}
    \caption{\small Lift at 0.5 of the Jewish demographic when talking about Judaism using Yiddish/Hebrew, as contrasted with 3 other demographics.}
    \label{fig:jewish-lift}
\end{figure}

\begin{figure}[h]
    \centering
    \begin{subfigure}[b]{\linewidth}
        \centering
        \includegraphics[width=\linewidth]{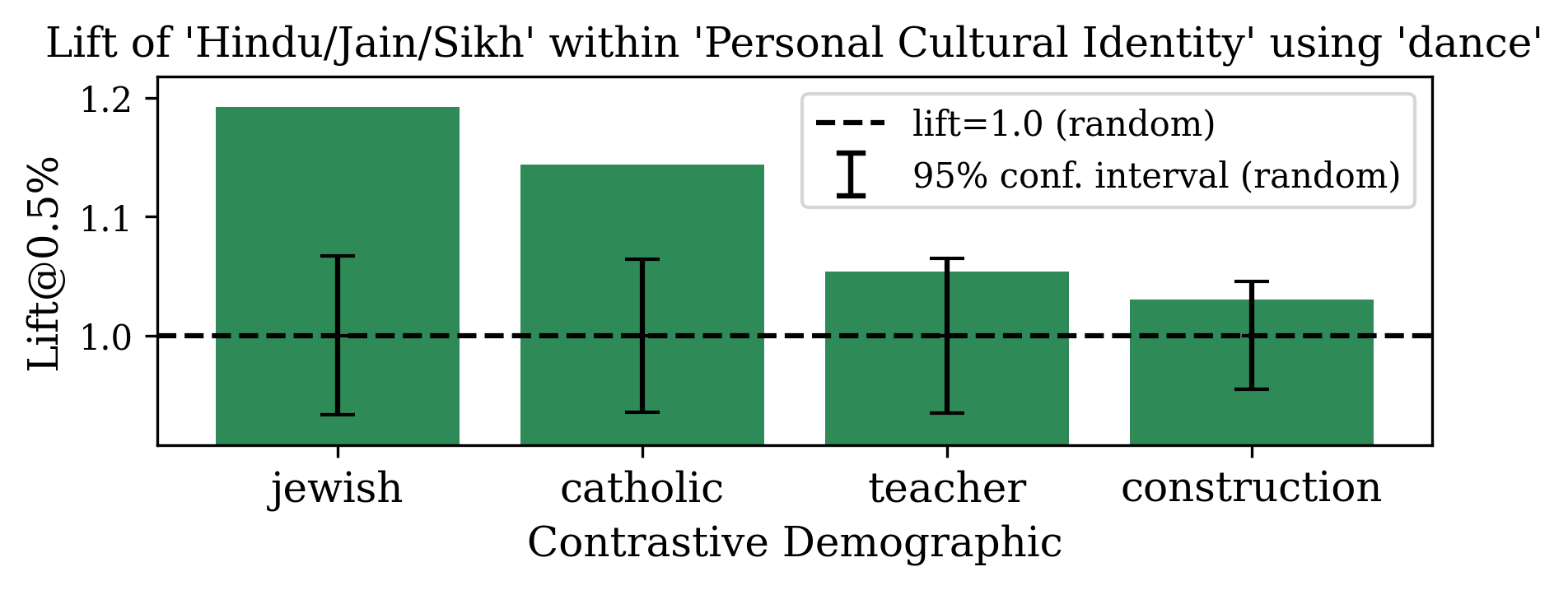}
    \end{subfigure}
    \caption{\small Lift at 0.5 of the Hindu Jain Sikh demographic when talking about Personal Cultural Identity using "dance", as contrasted with 4 other demographics.}
    \label{fig:dance-lifts}
\end{figure}

Jewish: (avg. $1.124 @0.5\%$), but are about average in triviality ($0.746$ Triviality)

\section{PSLP Automatic Evaluation}
\label{app:pslp-automatic-evaluation}
Table~\ref{tab:demographic_lexica} shows the lexica $\ell$ that were used to compute the Triviality for each demographic.

We used an LLM to generate over 23,000 candidate PSLPs. We used five prompt variations that gave the model increasing levels of context: from just the target demographic ($A$), to the demographic pair ($A,B$), to the full context including the topic ($t$). A final ``creative'' prompt also used the full context but was instructed to generate more novel lexica (all prompts are in App.~\ref{app:prompts}).

\begin{figure}[h]
    \centering
    \begin{subfigure}[b]{\linewidth}
        \centering
        \includegraphics[width=\linewidth]{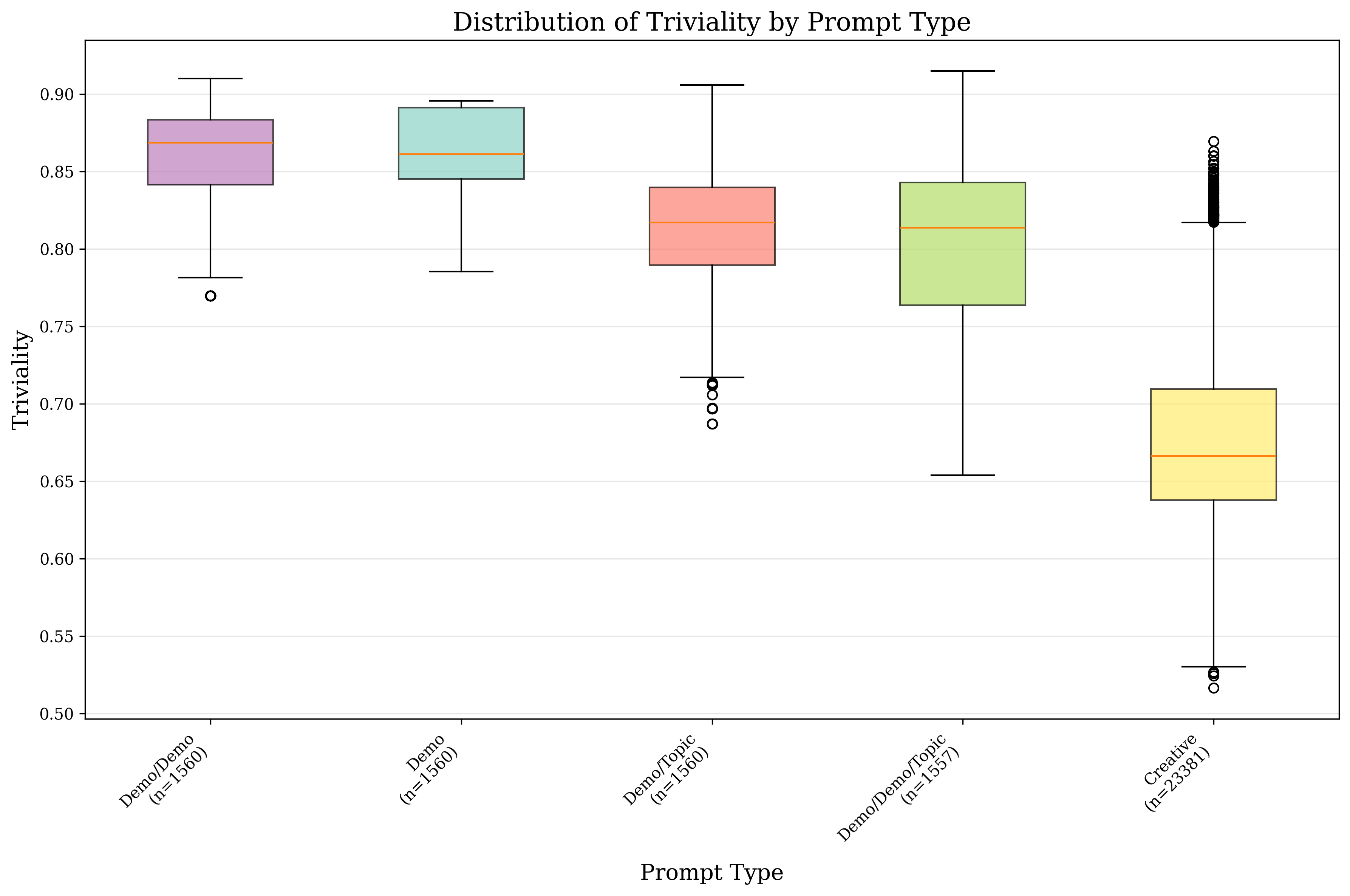}
    \end{subfigure}
    \caption{\small Distributions of Triviality by prompt type.}
    \label{fig:triviality-vs-prompt}
\end{figure}

We see the difference in triviality induced by the prompt: excluding the topic (as in the Demo and Demo/Demo prompts) gives more trivial PSLPs, while the `Creative' prompt gives the most nontrivial PSLPs. The distribution of each prompt's triviality (without lift) can be found in Figure~\ref{fig:triviality-vs-prompt}, and lift can be found in Table~\ref{tab:regex-vs-regex-llm}.

\begin{table}[ht]
\tiny
\centering
\begin{tabular}{l l}
\hline
\textbf{Demographic} & \textbf{Lexicon~($\ell$)} \\
\hline
Teacher & \makecell[l]{teacher, educator, education, \\ teaching, teach, schoolteacher} \\[0.7em]
Catholic & \makecell[l]{Catholic, Catholicism, Catholic Church, \\ Mass, Eucharist, Catechism, Catholic priest} \\[0.7em]
Black & \makecell[l]{Black, African American, Black history, \\ Afro‑American, Black people, Black person} \\[0.7em]
\makecell[l]{Construction\\\hspace{1em}Worker} 
        & \makecell[l]{construction worker, construction, builder, \\ construction site, contractor, building, laborer} \\[0.7em]
Jewish  & \makecell[l]{Jewish, Jew, Judaism, Jewish holidays, \\ Torah, synagogue, Kosher, Shabbat, Rabbi} \\[0.7em]
\makecell[l]{Hindu/Jain/\\\hspace{1em}Sikh} 
        & \makecell[l]{Hindu, Hinduism, Jain, Jainism, Sikh, Sikhism, \\ puja, Shiva, Vishnu, ahimsa, karma, Gurdwara} \\
\hline
\end{tabular}
\caption{Demographic lexica for triviality calculation.}
\label{tab:demographic_lexica}
\end{table}

\begin{table}[h!]
\small
\centering
\begin{tabular}{l ccccc}
\hline
\textbf{Prompt Type} & \textbf{mean} & \textbf{min} & \textbf{q25} & \textbf{q75} & \textbf{max} \\
\hline
D1 + D2 & 1.28 & 0.43 & 1.05 & 1.37 & 6.84 \\
\shortstack{D1 + D2 +\\Topic} & 1.32 & 0.25 & 1.03 & 1.40 & 7.78 \\
D1 + Topic & 1.20 & 0.11 & 1.00 & 1.29 & 5.09 \\
D1 & 1.44 & 0.45 & 1.11 & 1.56 & 7.67 \\
\shortstack{D1 + D2 +\\Topic (Theory)} & 1.07 & 0.05 & 0.96 & 1.13 & 6.71 \\
\hline
\end{tabular}
\caption{Distribution of lift@0.5 by prompt type.}
\label{tab:lift-vs-prompt}
\end{table}

\section{Human Validation of the Triviality Metric}
\label{app:human-validation-app}

This section provides a detailed account of the human validation study conducted to assess the effectiveness of our Triviality metric as a proxy for human judgments of ``unexpectedness.'' The study involved 9 annotators who collectively spent approximately 7 hours on the task. The annotators were from diverse demographic backgrounds, including members of 3 of the 6 groups studied in our dataset, and all held at least a Bachelor's degree with fluency in English. After filtering for significant lift PSLPs (p < 0.025), we randomly sampled 500 PSLPs for this study, with each PSLP being rated by 3 of the 9 annotators to ensure reliable judgments.

The annotation task was designed to elicit an intuitive judgment of surprise. For each instance, annotators were presented with a Target Demographic (e.g., ``Jewish''), a Contrast Demographic (e.g., "Catholic"), a Topic (e.g., "Elections"), and a Lexicon (e.g., `{"ballot access", "voter registration", "gerrymandering"}`). They were then instructed to rate the lexicon based on the following instructions:

\emph{1. First think about what you know about the two demographics A and B, especially when they talk about the given topic. What kinds of words/phrases might Demographic A use that Demographic B would not? Specifically, we care about such words/phrases that are not obvious, or unexpected.\\Example: When talking about "recipes", Indian people when contrasted with American people may use words (ingredients) that are common in Indian recipes (cardamom or saffron, turmeric, etc.)\\Note: When Indian and Bangladeshi people are contrasted, this might not work because such words are shared across the two groups.}

\emph{2. Once you are ready, think about how these keywords compare to what you came up with. Were you surprised that the keywords worked in distinguishing the two groups?}


To measure the consistency and quality of the annotations, we calculated the Intraclass Correlation Coefficient. Using the `ICC(2,k)' two-way random effects model, which assesses the reliability of the average ratings, we achieved a score of \textbf{0.738}. This indicates ``good'' reliability and confirms that annotators shared a consistent understanding of the task.

Our analysis yielded two key results. First, we computed the Spearman rank correlation ($\rho$) between our automated Triviality score and the average human unexpectedness score for all 500 PSLPs, finding a significant negative correlation of $\rho = -0.38$ ($p < 0.001$). This supports our hypothesis that as a PSLP becomes more trivial according to our metric, it is perceived as less unexpected by humans. Second, we evaluated the metric's utility as a filter. We defined a PSLP as ``promising'' if all 3 annotators assigned it an unexpectedness score of 3 or higher, which applied to 135 of the 500 instances (27\%). The \textbf{1.5-1.8x improvement factor} reported in the main paper represents the reduction in manual labor; it is the ratio of PSLPs one must inspect to find a ``Promising'' one without the filter versus with the filter. For example, if a researcher must review 100 statistically significant PSLPs to find 10 ``Promising'' ones (a 10:1 effort), and our Triviality filter narrows the pool to 60 candidates that still contain 9 of the ``Promising'' ones (a 6.7:1 effort), the reduction in effort is $10 / 6.7 \approx 1.5x$. The 1.5-1.8x range reflects this practical speedup at reasonable operating thresholds for the filter.
\section{Prompts}
\label{app:prompts}

All LLM calls were done using gpt-4.1-nano-2025-04-14, except for PSLP generation, where we used gpt-4.1-2025-04-14.

\begin{figure}[h!]
    \centering
    \begin{tcolorbox}[colback=gray!30, colframe=black, boxrule=0.5mm, width=\columnwidth]
    \tiny
    \#\#\# Task Overview:\\
    As a social media analysis assistant, your task is to analyze a social media post and determine if the user has self-identified themselves. You will be given a post, and a target demographic (e.g. "Black", "Teacher", etc). Your task is to read the post and determine with high confidence whether the user has self-identified themselves as the demographic (e.g. "I am a black man"). In addition, you must determine whether the user has self-identified as a demographic that is mutually exclusive to the target demographic (e.g., for "black", this could be saying "I am a white woman" or "I am not black"; for "teacher", this could be saying "I work in construction" or "I am not a teacher").\\
    
    \#\#\# Response Format:\\
    Your response must adhere to the format below:\\
    User self-identifies as demographic: yes OR no\\
    User self-identifies as mutually exclusive demographic: yes OR no\\
    
    \#\#\# Demographic\\
    \{demographic\}\\
    
    \#\#\# Social Media Post\\
    \{post\}\\
    
    \#\#\# Response
    \end{tcolorbox}
    \caption{Prompt for self-id or anti-self-id.}
    \label{fig:self-id}
\end{figure}

\begin{figure}[h!]
    \centering
    \begin{tcolorbox}[colback=gray!30, colframe=black, boxrule=0.5mm, width=\columnwidth]
    \tiny
    \#\#\# Task Overview:\\
    You are a socio-linguistic scientist. You will answer questions posed by the user, taking into consideration every detail of their request. Format the output in the same way as the example provided.\\

    \#\#\# Task Details:\\
    You are given two demographics; a \textbf{target demographic} and a \textbf{contrast demographic}. Your task is to generate words and phrases that would be used in posts by the target demographic but are unlikely to be used by the contrast demographic. The posts will be mostly in the English script.\\

    \#\#\# Strict Rules:\\
    1. \textbf{ONLY} output words (and phrases) that can be directly searched. No theories like ``they might use...''\\
    2. Output as many words and phrases as you think are appropriate:\\
    \quad -- If there are many differentiating things, output many words. Otherwise output fewer but \textbf{high quality} words.\\
    3. First think about the demographics given out loud. Mention out loud what you know about them and then generate the words.\\
    4. \textbf{Don't} output reasoning with the words, just the words and phrases.\\

    \#\#\# Example Input:\\
    Target demographic: Chinese\\
    Contrast demographic: Russian\\

    \#\#\# Example Output:\\
    The \textbf{target demographic} is Chinese, and the \textbf{contrast demographic} is Russian. The goal is to find \textbf{English-language} words and phrases that appear in posts by people from the Chinese demographic but are \textbf{unlikely} to appear in posts by Russians. The words must be \textbf{searchable}, i.e., things they would actually write or say online, in English. People from mainland China or ethnic Chinese individuals posting in English may exhibit certain linguistic, cultural, or stylistic markers that set them apart from Russian posters.\\

    Key distinguishing features to think about:\\
    * \textbf{Cultural references}: Chinese users may refer to cultural institutions, apps, celebrities, or ideologies unique to China.\\
    * \textbf{Language influence}: Certain phrasings that come from direct translation of Chinese idioms or grammar.\\
    * \textbf{Internet platforms}: Use of uniquely Chinese platforms like WeChat, Weibo, or Bilibili.\\
    * \textbf{Education/study culture}: Common English phrases around Gaokao, overseas study, etc.\\
    * \textbf{Political language}: Specific phrasing from Chinese state media, such as ``core socialist values'' or ``Western hostile forces'', not likely to be used by Russians.\\

    Words and Phrases:\\
    1. WeChat\\
    2. Douyin\\
    3. Gaokao\\
    4. 996 work culture\\
    5. Spring Festival Gala\\
    6. Bilibili\\
    7. Little Red Book\\
    8. Weibo\\
    9. Tsinghua University\\
    10. C-pop\\
    ...\\

    \#\#\# Task Input:\\
    Target demographic: \{target\}\\
    Contrast demographic: \{contrast\}\\

    \#\#\# Task Output:
    \end{tcolorbox}
    \caption{Prompt when 2 demographics are given.}
    \label{fig:demo-demo-prompt}
\end{figure}

\begin{figure}[h!]
    \centering
    \begin{tcolorbox}[colback=gray!30, colframe=black, boxrule=0.5mm, width=\columnwidth]
    \tiny
    \#\#\# Task Overview:\\
    You are a socio-linguistic scientist. You will answer questions posed by the user, taking into consideration every detail of their request. Format the output in the same way as the example provided.\\

    \#\#\# Task Details:\\
    You are given a \textbf{target demographic}. Your task is to generate words and phrases that would be used in posts by the target demographic. The words and phrases should be relatively unique to this demographic so that we can identify documents authored by this demographic. The posts will be mostly in the English script.\\

    \#\#\# Strict Rules:\\
    1. \textbf{ONLY} output words (and phrases) that can be directly searched. No theories like "they might use..."\\
    2. Output as many words and phrases as you think are appropriate:\\
    \quad - If there are many differentiating things, output many words. Otherwise output less but \textbf{high quality} words.\\
    3. First think about the demographics given out loud. Mention out loud what you know about them and then generate the words.\\
    4. Don't output reasoning with the words, just the words and phrases.\\

    \#\#\# Example Input:\\
    Target demographic: Sri Lankan\\

    \#\#\# Example Output:\\
    Sri Lankans are a South Asian demographic with diverse linguistic and cultural backgrounds, primarily Sinhala and Tamil speakers, but English is widely used in online posts, especially among urban youth and diaspora communities. Common topics include cricket, politics, local food, and religious/cultural festivals like Vesak, Sinhala and Tamil New Year, and Poson. There’s also heavy usage of local slang, transliterated Sinhala/Tamil phrases, and references to places, politicians, and celebrities. English posts by Sri Lankans often include code-switching (Sinhala/Tamil-English hybrid sentences), slang like “machan” (buddy), and references to foods like kottu, hoppers, and short eats. The political discourse often includes names of local parties (SLPP, UNP, JVP), figures (Gotabaya, Mahinda, Ranil), and issues like “aragalaya” (protest movement). Diaspora posts may talk about missing Sri Lanka, family back home, and visiting “lanka” during holidays.\\

    Words and Phrases:\\
    1. machan\\
    2. kottu\\
    3. hoppers\\
    4. short eats\\
    5. poya day\\
    6. Vesak\\
    7. Avurudu\\
    8. Poson\\
    9. parippu\\
    10. rice and curry\\
    11. dhal curry\\
    12. sambol\\
    13. thambili\\
    14. aragalaya\\
    15. Gota\\
    16. Mahinda\\
    17. Ranil\\
    ...\\

    \#\#\# Task Input:\\
    Target demographic: \{target\}\\

    \#\#\# Task Output:
    \end{tcolorbox}
    \caption{Prompt when only target demographics is given.}
    \label{fig:demo-prompt}
\end{figure}

\begin{figure}[h!]
    \centering
    \begin{tcolorbox}[colback=gray!30, colframe=black, boxrule=0.5mm, width=\columnwidth]
    \tiny
    \#\#\# Task Overview:\\
    You are a socio-linguistic scientist. You will answer questions posed by the user, taking into consideration every detail of their request. Format the output in the same way as the example provided.\\
    
    \#\#\# Task Details:\\
    You are given a \textbf{topic} and a \textbf{target demographic}. Your task is to generate words and phrases that are likely to be used in posts by the target demographic when talking about the given topic. The words and phrases should be such that they help identify the demographic as the author of documents about the topic. The posts will be mostly in the English script.\\
    
    \#\#\# Strict Rules:\\
    1. ONLY output words (and phrases) that can be directly searched. No theories like ``they might use...''\\
    2. Output as many words and phrases as you think are appropriate:\\
    \hspace*{2em}- If there are many differentiating things, output many words. Otherwise output less but \textbf{high quality} words.\\
    3. First think about the demographics given out loud. Mention out loud what you know about them and then generate the words.\\
    4. Don't output reasoning with the words, just the words and phrases.\\

    \#\#\# Example Input:\\
    Target demographic: Sri Lankan\\
    Topic: cricket\\

    \#\#\# Example Output:\\
    Sri Lankans are deeply passionate about cricket—it's the most popular sport in the country and a major source of national pride. Sri Lankan cricket fans often reference legendary players, local teams, and emotional highs and lows of key matches, especially against rivals like India, Pakistan, and Australia. They use Sinhala or Tamil terms occasionally, but in English posts, they tend to blend cricket lingo with cultural pride. They also often refer to specific tournaments (e.g., Asia Cup, World Cup), iconic moments, and local slang or nicknames for players. There is a strong tendency to mention players from different eras (e.g., Sanath, Sanga, Mahela, Malinga, etc.) and to express strong emotions about wins, losses, and team selections.\\
    \#\\
    Words and Phrases:\\
    1. Lion pride\\
    2. Lankan Lions\\
    3. Mahela magic\\
    4. Sanga class\\
    5. Sanath sixes\\
    6. Malinga yorkers\\
    7. Dilshan scoop\\
    8. Kusal form\\
    9. Lankan fighting spirit\\
    10. Island nation roar\\
    11. SL cricket\\
    12. Come on Sri Lanka\\
    13. Lankan powerplay\\
    14. Proud to be Sri Lankan\\
    ...\\

    \#\#\# Task Input:\\
    Target demographic: \{target\}\\
    Topic: \{topic\}\\

    \#\#\# Task Output
    \end{tcolorbox}
    \caption{Prompt when target demographic and topic is given.}
    \label{fig:demo-topic-prompt}
\end{figure}

\begin{figure}[h!]
    \centering
    \begin{tcolorbox}[colback=gray!30, colframe=black, boxrule=0.5mm, width=\columnwidth]
    \tiny
    \#\#\# Task Overview:\\
    You are a socio-linguistic scientist. You will answer questions posed by the user, taking into consideration every detail of their request. Format the output in the same way as the example provided.\\
    
    \#\#\# Task Details:\\
    You are given a \textbf{topic} and two demographics; a \textbf{target demographic} and a \textbf{contrast demographic}. Your task is to generate words and phrases that are likely to be used in posts by the target demographic but unlikely to be used by the contrast demographic, for the given topic. The posts will be mostly in the English script.\\
    
    \#\#\# Strict Rules:\\
    1. \textbf{ONLY} output words (and phrases) that can be directly searched. No theories like "they might use..."\\
    2. Output as many words and phrases as you think are appropriate:\\
    \quad - If there are many differentiating things, output many words. Otherwise output less but \textbf{high quality} words.\\
    3. First think about the demographics given \textbf{out loud}. Mention out loud what you know about them and then generate the words.\\
    4. Don't output reasoning with the words, just the words and phrases.\\
    
    \#\#\# Example Input:\\
    Target demographic: Chinese\\
    Contrast demographic: Russian\\
    Topic: cooking\\
    
    \#\#\# Example Output:\\
    \textbf{Reasoning:}\\
    Chinese cooking culture emphasizes diverse regional cuisines like Sichuan, Cantonese, Hunan, and Shanghainese. It includes techniques such as stir-frying, steaming, braising, and focuses on ingredients like soy sauce, ginger, garlic, and fermented products (e.g., doubanjiang, fermented tofu). Terms are often borrowed from Chinese languages even when writing in English, particularly for dish names, ingredients, and cooking methods. Russian cooking, on the other hand, leans toward root vegetables, dairy, baking, and stewing. There’s less focus on wok cooking or umami-rich fermented ingredients. While both cultures value home cooking, the vocabulary around the ingredients and cooking techniques differs significantly. Therefore, we can identify words and phrases commonly used in English-language posts by Chinese individuals that reflect unique elements of Chinese culinary tradition, which are much less likely to appear in Russian cooking posts.\\
    Words and Phrases:\\
    1. wok hei\\
    2. doubanjiang\\
    3. red braised pork\\
    4. mala hotpot\\
    5. liangpi\\
    \ldots\\
    
    \#\#\# Task Input:\\
    Target demographic: \{target\}\\
    Contrast demographic: \{contrast\}\\
    Topic: \{topic\}\\
    
    \#\#\# Task Output
    \end{tcolorbox}
    \caption{Prompt when both demographics and topic is given.}
    \label{fig:socioling-prompt}
\end{figure}

\begin{figure}[h!]
    \centering
    \begin{tcolorbox}[colback=gray!30, colframe=black, boxrule=0.5mm, width=\columnwidth]
    \tiny
    \#\#\# Task Details:\\
    You are given a \textbf{topic} and two demographics; a \textbf{target demographic} and \textbf{contrast demographic}. Your task is to come up with \textbf{15 cultural, sociological, or linguistic theories} about how the \textbf{target group} talks about the topic, especially as opposed to the contrast demographic. Then for each theory, come up with keywords and phrases to help retrieve posts from the target demographic. Output at least 10 words for each theory. Follow the response format specified below.\\
    
    \#\#\# Instruction for Theories:\\
    - The theories should be \textbf{nuanced and sophisticated}. Don't just combine demographic + topic.\\
    \ \ \ \ \textemdash{} Some bad examples for the topic ``Bread'':\\
    \ \ \ \ \ \ \ \ $\cdot$ Hindus eat naan, roti, paratha, etc.\\
    \ \ \ \ \ \ \ \ $\cdot$ Catholics talk about eucharist, host, communion, etc.\\
    - Start by thinking about the factors that are relevant to the topic (which is not demographic dependent).\\
    \ \ \ \ \textemdash{} For example, for the topic ``Health'', the factors that would be relevant are genetics, age, diet, physical activity, substance use, healthcare access, financial status, etc.\\
    - Then think about how these factors manifest in the target demographic, especially as opposed to the contrast demographic.\\
    \ \ \ \ \textemdash{} For example, for the factor ``genetics'', target demographic ``Hispanics'', and contrast demographic ``Black'':\\
    \ \ \ \ \ \ \ \ Good example: gallbladder disease (considered high predisposition in Hispanics but low in Black people).\\
    \ \ \ \ \ \ \ \ Bad example: Type 2 Diabetes (both groups have high predisposition to this).\\
    - The theories will be validated by real sociolinguists and social scientists, so try to come up with new theories that most people would not think of. Don't be conservative, instead get creative!\\

    \#\#\# Instruction for Keywords and Phrases:\\
    - Each word/phrase will be used for a direct look-up so focus on high quality words.\\
    - Make sure each word/phrase is likely to be used a lot.\\
    \ \ \ \ \textemdash{} Bad example: ``spirit flowing through the bones''\\
    \ \ \ \ \textemdash{} Good example: ``warm tonic''\\
    - Don't use words or phrases that could only ever be used by one demographic (e.g. ``ahimsa'', ``moksha'', ``kosher'', ``mitzvah'', ``catechesis'', ``sacrament'', etc.). This includes demographic-specific foreign language words, holidays, traditions, etc. Do not use these words.\\

    \#\#\# Response Format (angle brackets are placeholders for your output):\\
    Reasoning: \textless The factors I think are relevant are... This is what I know about this demographic related to these factors... contrasted with the contrast demographic I think... therefore....\textgreater\\

    Final Outputs:\\
    1. Theory 1: \textless your first theory\textgreater\\
    Keywords and Phrases: \textless word\textgreater, \textless phrase\textgreater, ...\\
    \vdots\\
    15. Theory 15: \textless your last theory\textgreater\\
    Keywords and Phrases: \textless word\textgreater, \textless phrase\textgreater, ...\\

    \#\#\# Input:\\
    Target demographic: \{target\}\\
    Contrast demographic: \{contrast\}\\
    Topic: \{topic\}
    \end{tcolorbox}
    \caption{Prompt when both demographics and topic is given to generate creative lexicon.}
    \label{fig:socio-theory}
\end{figure}

\begin{figure}[h!]
    \centering
    \begin{tcolorbox}[colback=gray!30, colframe=black, boxrule=0.5mm, width=\columnwidth]
    \tiny
    \#\#\# Task Details:\\
    You are given a research paper that studied a **target demographic** in a certain **context**. Your task is to extract all **cultural, sociological, or linguistic findings** that were demonstrated in the paper. Then, as a direct consequence of each finding being true, come up with 3 sets of keywords and phrases that would be used by target demographic within that context. Output at least 5 words/phrases in each set. Follow the response format specified below.\\

    \#\#\# Instruction for Findings:\\
    - The findings should be direct results demonstrated in the paper.\\
    
    \#\#\# Instruction for Keywords and Phrases:\\
    - Given that you know the findings are proven, what set of keywords/phrases would be used by the target demographic more than other demographics within the context?\\
    - Each word/phrase will be used for a direct look-up so focus on high quality words.\\
    - Make sure each word/phrase is likely to be used a lot.\\
        - Bad examples: *spirit flowing through the bones*, *it felt like such a long day*, *wasn't even making eye contact with me*\\
        - Good examples: *warm tonic*, *haven*, *supported*, *worth exploring*\\
    - Don't use words or phrases that could only ever be used by one demographic (e.g. *ahimsa*, *moksha* (Hindus), *kosher*, *mitzvah* (Jewish people), *catechesis*, *sacrament*, (Catholics) etc.). This includes demographic-specific foreign language words, holidays, traditions, etc. Do not use these words.\\
    
    \#\#\# Response Format (angle brackets are placeholders for your output):\\
    1. Finding: <extracted finding>\\
    	- Keywords and Phrases: <word>, <phrase>, ... (at least 5)\\
    	- Keywords and Phrases: <word>, <phrase>, ... (at least 5)\\
    	- Keywords and Phrases: <word>, <phrase>, ... (at least 5)\\
    2. Finding: <extracted finding>\\
    	- Keywords and Phrases: <word>, <phrase>, ... (at least 5)\\
    	- Keywords and Phrases: <word>, <phrase>, ... (at least 5)\\
    	- Keywords and Phrases: <word>, <phrase>, ... (at least 5)\\
    ...
    
    \#\#\# Input:
    Target demographic: \{target\}\\
    Context: \{topic\}\\
    Paper: (attached)
    \end{tcolorbox}
    \caption{Prompt to generate a lexicon aligning with the results of a published research paper.}
    \label{fig:lit-inspired-pslp-prompt}
\end{figure}

\section{Literature-Inspired PSLPs}
\label{app:lit-inspired-pslps}

Table~\ref{tab:lit-inspired-pslps} contains the 11 demographic/topic pairs and academic papers which were used to generate the Literature-Inspired PSLPs.

\subsection{Human Validation of Literature-Inspired PSLPs' Alignment with Papers}
\label{app:human-val-lit-pslp}
When creating the Literature-inspired PSLPs, the LLM was conditioned on published social science papers. Despite this conditioning to produce the lexica, we wanted to test whether the generated lexica truly aligned with the results of the paper. To do this, we had a human annotator inspect 44 of the Literature-inspired lexica. For each one, we showed the topic, demographic, and the lexicon, mixed with 3 other lexica for the that topic/demographic that were \textit{not} conditioned on the paper. The annotator was allowed to view the relevant paper, and was asked to pick which lexicon of the 4 truly came from the results of the paper. Of the 44, the annotator picked correctly 42 times (95.45\%), showing that in some way, the Literature-inspired PSLPs were truly inspired by the findings of their respective papers.

\begin{table*}[tb]
  \centering
  \setlength{\tabcolsep}{3pt}
  \resizebox{\textwidth}{!}{
  \begin{tabular}{llll}
    \toprule
    \textbf{Demographic} & \textbf{Topic} & \textbf{Lexica (combined, sampled)} & \textbf{Paper} \\
    \midrule
    \multirow{14}{*}{\textbf{Black}} 
      & \multirow{5}{*}{\shortstack[l]{Cooking/\\Baking}} 
        & pork chops, & \multirow{5}{*}{\shortstack[l]{Cultural aspects of African\\American eating patterns \cite{airhihenbuwa_cultural_1996}}}\\
      & & family dinner table, eat at home,\\
      & & soul food, adapting recipes,\\
      & & switching ingredients, breaking bread\\
      & & together, Sunday dinner \\
    \cmidrule(lr){2-4}
      & \multirow{3}{*}{\shortstack[l]{Healthcare}}
        & didn't listen, talked over me, & \multirow{3}{*}{\shortstack[l]{African American experiences in healthcare:\\``I always feel like I'm getting skipped over'' \cite{cuevas_ag_african_2016}}}\\
      & & second opinion, questioning everything,\\
      & & really listens, understands me \\
    \cmidrule(lr){2-4}
      & \multirow{6}{*}{\shortstack[l]{Metal/\\Rock\\Music}}
        & something else, don't fit in, & \multirow{6}{*}{\shortstack[l]{Black Metal Soul Music: Stone Vengeance\\and the Aesthetics of Race in Heavy Metal \cite{fellezs_black_2012}}}\\
      & & always a statement,\\
      & & the source,\\
      & & the right look, don't fit the image,\\
      & & technical skill, prove myself, safe to like,\\
      & & empty praise \\
    \midrule
    \multirow{13}{*}{\textbf{Jewish}} 
      & \multirow{4}{*}{\shortstack[l]{Superheroes/\\Comic Book\\Media}}
        & secret identity, living two lives, & \multirow{4}{*}{\shortstack[l]{Public Heroes, Secret Jews:\\Jewish Identity and Comic Books \cite{caplan_public_2021}}}\\
      & & finally revealed, true self,\\
      & & feeling like a monster, outsider status,\\
      & & tragic backstory, shaped by his past \\
    \cmidrule(lr){2-4}
      & \multirow{5}{*}{\shortstack[l]{Soccer}}
        & badge of honor, form of solidarity, & \multirow{5}{*}{\shortstack[l]{The Making of ``Jew Clubs'': Performing\\Jewishness and Antisemitism in European\\Soccer and Fan Cultures \cite{brunssen_making_2023}}}\\
      & & feel comfortable,\\
      & & remembering\\
      & & our history, in contrast to them, what\\
      & & separates us,\\
    \cmidrule(lr){2-4}
      & \multirow{4}{*}{\shortstack[l]{Hip-Hop/\\Rap Music}}
        & prove myself, not like that, & \multirow{4}{*}{\shortstack[l]{Jewish Flow: Performing Identity\\in Hip-Hop Music \cite{stein_jewish_2019}}}\\
      & & same struggle, our histories,\\
      & & out of place, I know it's weird,\\
      & & new path, found my way \\
    \midrule
    \multirow{5}{*}{\textbf{Catholic}} 
      & \multirow{2}{*}{\shortstack[l]{Video\\Games}}
        & moral compass, personal conviction, & \multirow{2}{*}{\shortstack[l]{Gaming Religionworlds: Why Religious Studies\\Should Pay Attention to Religion in Gaming \cite{campbell_gaming_2016}}}\\
      & & fighting darkness, a greater good \\
    \cmidrule(lr){2-4}
      & \multirow{3}{*}{\shortstack[l]{Banking \&\\Financial\\Institutions}}
        & stewardship, social responsibility, & \multirow{3}{*}{\shortstack[l]{Application of Catholic Social Teaching\\in Finance and Management \cite{czerwonka_application_2024}}}\\
      & & moral screening,\\
      & & serving the team, people first \\
    \midrule
    \multirow{5}{*}{\textbf{Hindu/Jain/Sikh}} 
      & \multirow{5}{*}{\shortstack[l]{Book/\\Literature\\Discussions}}
        & political dynasty, & \multirow{5}{*}{\shortstack[l]{Reinterpretation of Hindu Myths in\\Contemporary Indian English Literature \cite{jain_reinterpretation_2016}}}\\
      & & her side of the story, a woman's\\
      & & perspective, a fresh take, a reluctant hero,\\
      & & our cultural heritage, passing down\\
      & & the stories \\
    \midrule
    \multirow{4}{*}{\textbf{Teacher}} 
      & \multirow{4}{*}{\shortstack[l]{Anime/\\Manga}}
        & gateway text, a good start, & \multirow{4}{*}{\shortstack[l]{Learning past the pictures in the panels:\\teacher attitudes to manga and anime texts \cite{cheung_learning_2015}}}\\
      & & visual literacy, film techniques,\\
      & & student-led discussion, the real experts,\\
      & & the manga kids, their little group \\
    \midrule
    \multirow{2}{*}{\textbf{Construction}} 
      & \multirow{2}{*}{\shortstack[l]{Public Sector/\\Government Jobs}}
        & a good job, honest living, got my back, & \multirow{2}{*}{\shortstack[l]{An examination of blue- versus white-collar\\workers' conceptualizations of job satisfaction facets \cite{hu_examination_2010}}}\\
      & & good crew, fair wage, pays the bills \\
    \bottomrule
  \end{tabular}
  }
  \caption{Literature-inspired PSLPs across demographics and topics.}
  \label{tab:lit-inspired-pslps}
\end{table*}

\end{document}